%% file: diversity.tex
\documentclass[10pt,twocolumn,letterpaper]{article}

\usepackage{iccv}
\usepackage{times}
\usepackage{epsfig}
\usepackage{graphicx}
\usepackage{amsmath}
\usepackage{amssymb}

\usepackage{multirow}
\usepackage{tabularx}
\usepackage{listings}
\usepackage{xcolor}
\usepackage{setspace}
\usepackage{float}
\lstdefinelanguage{json}{
  basicstyle=\small\ttfamily,
  numbers=left,
  numberstyle=\tiny,
  stepnumber=1,
  numbersep=8pt,
  showstringspaces=false,
  breaklines=true,
  frame=lines,
  backgroundcolor=\color{gray!10},
  morestring=[b]{"},
  stringstyle=\color{blue},
  commentstyle=\color{red},
  keywordstyle=\color{cyan},
  morekeywords={true,false,null}
}

\usepackage{ifthen}
\usepackage[export]{adjustbox}
\usepackage{subcaption,graphicx}

\usepackage[breaklinks=true,bookmarks=false]{hyperref}

\iccvfinalcopy 

\def\iccvPaperID{****} 

\ificcvfinal\fi

\begin{document}

\title{\input{title.tex}}
\input{authors.tex}

\maketitle
\ificcvfinal\thispagestyle{empty}\fi

\newcommand{\imagenetcbar}{ImageNet-$\overline{\mbox{C}}$ }
\newcommand{\mrq}[1]{\textbf{#1}}
\newcommand{\rqspace}{}

\begin{abstract}
\input{abstract.tex}
\end{abstract}
\input{content2.tex}

\paragraph{Acknowledgements}

This work was partially funded by the Federal Ministry for Economic Affairs and Climate Action of Germany, as part of the research project SafeWahr (Grant Number: 19A21026C).

We would also like to thank Professors Lorenzo Strigini and Peter Popov for the fruitful conversations and feedback on this work.

{\small
\bibliographystyle{ieee_fullname}
\bibliography{egbib}
}

\input{suplemental.tex}

\end{document}

%% file: title.tex
Exploring Resiliency to Natural Image Corruptions in Deep Learning using Design Diversity

%% file: authors.tex
\author{Rafael Rosales, Pablo Munoz, Michael Paulitsch\\
Intel Labs, Intel Corporation\\
{\tt\small \{rafael.rosales, pablo.munoz, michael.paulitsch\}@intel.com}
}

%% file: abstract.tex
In this paper, we investigate the relationship between diversity metrics, accuracy, and resiliency to natural image corruptions of Deep Learning (DL) image classifier ensembles.
We investigate the potential of an attribution-based diversity metric to improve the known accuracy-diversity trade-off of the typical prediction-based diversity.
Our motivation is based on analytical studies of design diversity that have shown that a reduction of common failure modes is possible if diversity of design choices is achieved.

Using ResNet50 as a comparison baseline, we evaluate the resiliency of multiple individual DL model architectures against dataset distribution shifts corresponding to natural image corruptions.
We compare ensembles created with diverse model architectures trained either independently or through a Neural Architecture Search technique and evaluate the correlation of prediction-based and attribution-based diversity to the final ensemble accuracy.
We evaluate a set of diversity enforcement heuristics based on negative correlation learning to assess the final ensemble resilience to natural image corruptions and inspect the resulting prediction, activation, and attribution diversity.

Our key observations are:
1) model architecture is more important for resiliency than model size or model accuracy,
2) attribution-based diversity is less negatively correlated to the ensemble accuracy than prediction-based diversity,
3) a balanced loss function of individual and ensemble accuracy creates more resilient ensembles for image natural corruptions,
4) architecture diversity produces more diversity in all explored diversity metrics: predictions, attributions, and activations.

%% file: content2.tex
\section{Introduction}

    In the context of Deep Learning (DL), it has been empirically discovered that the use of ensembles can improve the model's accuracy in tasks such as regression and classification.
    It has been speculated~\cite{deepEnsemLoss} that the main reason behind these improvements is the implicit diversity in the solutions found that when aggregated as an ensemble obtain better predictions.
    In this work, we evaluate the resiliency of diverse deep learning classifiers and the role that the different kinds of diversity play in improving them.

    \subsection{The case for design diversity}
    	
        Design diversity~\cite{littlewood2001modeling} is a technique to increase the resilience of safety critical systems.
        It is established as a best practice in standards such as in vehicle functional safety ~\cite{ISO26262} to prevent dependent failures, safety of the intended functionality ~\cite{SOTIF} to address system limitations of machine-learning-based components, and avionic software~\cite{euroCAE}.
    
        A common pitfall is to misunderstand \textit{independent development} as \textit{design diversity}.
        In~\cite{boeing777} the designers of a safety-critical system preferred to let multiple teams collaborate, although the purpose of having multiple teams is to produce multiple designs of a single specification.
        This was justified with the claim that specification problems can be better mitigated with such collaboration but at the cost of the sought \textit{independence}.
    
        The key problem is, that independent development can (and will) produce designs with common failures mainly due to the fact that independent designs do not enforce diverse design choices.
        In fact, it has been statistically proven that independently developed software results in dependent failure behavior on randomly selected inputs~\cite{ELmodel}.
    
        In~\cite{LMmodel}, it has been shown that what is needed to reduce dependent failure behavior is diversity in design \textit{choices}.
        If the choices are made satisfying certain properties, it can be expected (in the average case) to obtain negatively correlated failure behavior, i.e., better than independent.
    
        In DL, however, design choices are not made by the human designers explicitly but are a result of the architecture, data, and optimization approach.
        Furthermore, existing diversity metrics are not directly related to the DL model's \textit{design choices} and are known to have a diversity-accuracy trade-off~\cite{KunchevaW03}.
        We investigate if a diversity metric closer to the model \textit{design choices} can improve the model resilience compared to existing metrics.
    
        \newcounter{reqno}
        \newcommand{\rrq}[2]{\textbf{RQ#1:}\refstepcounter{reqno}\label{req:#2}}
        \newcommand{\refreq}[1]{\textbf{RQ\ref{req:#1}}}

    \subsection{Main research questions}
    
        We aim to answer the following research questions:\\
        \rrq{1}{accRes} Is model accuracy, size, or architecture the main explanatory factor of resilience against natural image corruptions?\\
        \rqspace
        \rrq{2}{att} Can a diversity metric closer to design choices improve the known accuracy-diversity trade-off?\\
        \rqspace
        \rrq{3}{bestEnf} Which diversity enforcement heuristic produces the most resilient models?\\
        \rqspace
        \rrq{4}{divNCL} How diverse are the predictions, activations, and input feature attributions of models created with a diversity enforcement learning approach?\\    
    
        The rest of the paper is structured as follows: Section~\ref{sota} provides a brief overview of the current state-of-the-art in diversity enforcement and measurement.
        Next, the methodology is stated in Section~\ref{labelDesignDiversity}.
        Our experiments are then presented in Section~\ref{labelExperiments}, followed by a discussion of the outcomes and conclusions in Section~\ref{labelDiscussion}.

\section{Related work}
\label{sota}

    \subsection{Ensemble creation techniques} 
    
        The most relevant techniques for ensemble creation are:
        a) Ensembles of independently trained models where diversity originates from the training process randomness, e.g., seed.
        Each ensemble member loss is a function notated as:
    
        \begin{equation}
        \label{eq:ind}
            l(h^{i},y)
        \end{equation}
        where $y$ is the ground-truth label, and $h^{i}$ is the output of the $i$th single ensemble member.
        \cite{ensResilience} presents an analysis of the resilience of independently trained ensembles.
        b) Bagging~\cite{breiman1996bagging}, reduces the variance of multiple models by averaging the outcomes of models created with different training data subsets.
        c) Boosting~\cite{schapire1999brief} sequentially trains models to reduce bias by sampling incorrectly classified inputs more often in the next model.
        d) Negative Correlation Learning (NCL)~\cite{LIU} trains models \textit{in parallel} with a shared penalty term in their loss function to enforce prediction diversity.
        Generalized NCL (GNCL)~\cite{gncl} proposes two extensions for NCL: i) a generalized loss function for each member:
        \begin{equation}
            \label{eq:gncl_i}
            \sum_{i=1}^{M}{l(h^{i},y)-\frac{\lambda}{2M}\sum_{i=1}^{M}{{d_{i}}^{T}\mathfrak{D}d_{i}}}
        \end{equation}
        where $M$ is the total number of members in the ensemble, $d$ is the difference $h^i-f$ with $f$ as the ensemble prediction, $\mathfrak{D}$ is the 2nd derivative of the loss function, and $\lambda$ is a weighting hyper-parameter.
        ii) An implicit enforcement of diversity by balancing the ensemble and the individual loss:
        \begin{equation}
            \label{eq:balanced}
            \frac{1}{M}\sum_{i=1}^{M}{\big(\lambda{l(f,y)} + \frac{1-\lambda}{M}\sum_{i=1}^{M}{l(h^{i},y)}\big)}
        \end{equation}
    
    \subsection{Diversity metrics in DL}
    
        There are many proposed metrics for diversity.
        \cite{BrownWHY05} presents a survey and taxonomy for diversity metrics and \cite{GongZH19} presents a survey of diversity for ML.
        In this work, we focus on behavior diversity metrics of a DL model.
        Input data diversity such as different modalities or implementation aspects such as the number of layers is not considered.
    
        \begin{figure}
    	\includegraphics[width=\linewidth]{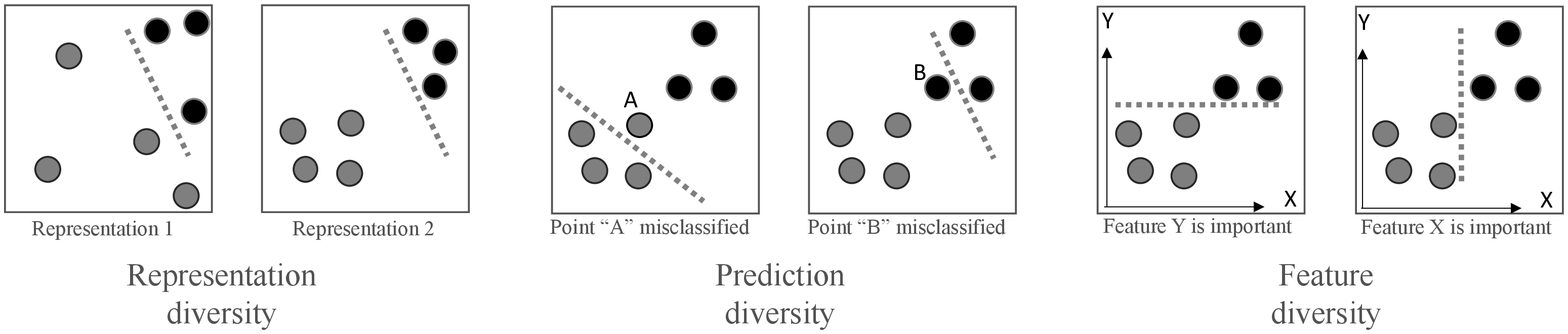}\caption{Model behavior diversity.
            a) Invariant decision boundary \& diverse sample representation.
            b) Diverse decision boundary and measured by prediction errors.
            c) Diverse decision boundary and measured by feature relevance. } \label{figDivTypes}
        \end{figure}
    
        \textbf{Prediction (output,  failures) diversity}
        Multiple prediction-based diversity metrics have been proposed. \cite{KunchevaW03} presents a comprehensive evaluation of this class of metrics. Pair-wise measures based on the correct and incorrect statistics of two models include the Q-statistics, correlation coefficient $\rho$, and the disagreement measure:
        \begin{equation}
            \label{eq:disagreement}
            D_{p,q}=\frac{N^{01}+N^{10}}{N^{11}+N^{10}+N^{01}+N^{00}}
        \end{equation}
        where the indexes $a,b$ of the binary vectors $N^{ab}$ indicate the correctness of the classifiers, i.e.,  $(h^{i=p}=y)\Rightarrow (a=1) \wedge (h^{i=p}\neq{y})\rightarrow(a=0)$.        
        Non-pairwise measures that evaluate non-binary diversity include entropy, coincident failure diversity, cosine similarity, Kullback–Leibler divergence~\cite{DvornikMS19,NamYLL21}, and the the Shannon equitability index~\cite{peet1975relative,ensResilience}:
        \begin{equation}
            \label{eq:shannon}
            E = -\sum_{i=1}^{S}{p_{i}\ln{p_{i}}}\big/\ln{S}
        \end{equation}
        where $S$ is the total number of prediction species/classes and $p_i$ is the proportion of observed species $i$.
    
        \textbf{Representation (activation) diversity}
        The intermediate representations (IR) can also be used to measure diversity.
        \cite{deepEnsemLoss} compares the diversity in representation space from independently created ensembles and ensembles from variational approaches.
        Measuring IR diversity is challenging due to space size and semantic ambiguity, i.e., the same semantic concept can be represented in many different ways.
        A naive use of any diversity metric such as cosine similarity could give semantically irrelevant diversity scores.
        In~\cite{Kornblith0LH19}, the Centered Kernel Alignment (CKA) metric is proposed to obtain a statistical measure across a dataset on the similarity of any two layers of a DL model:
        \begin{equation}
            \label{eq:cka}
            \text{CKA}(K,L) = \frac{\text{HSIC}(K,L)}{\sqrt{\text{HSIC}(K,K)\text{HSIC}(L,L)}}
        \end{equation}
        where $K$ and $L$ are similarity matrices of the two feature maps being compared and HSIC is the Hilbert-Schmidt Independence Criterion which measures statistical independence.
        The feature maps may be layer activations or attention maps such as Saliency, Integrated Gradients, and Grad-CAM~\cite{SimonyanVZ13, SelvarajuCDVPB17, SundararajanTY17}.
        
        \cite{QiKL21} proposed the use of the pull-away loss term from generative adversarial networks to induce diversity of such activations.
        Self-attention~\cite{VaswaniSPUJGKP17} (not related to attention maps) is one of the key techniques in the transformer architecture.
        In~\cite{pacmac} the embeddings used to feed attention heads are masked in such a way as to enforce diversity of activations.
        In zero-shot learning, the \textit{attribute} concept is used to enable training models that can, later on, predict unseen labels.
        These attributes can be considered for IR diversity, as well ~\cite{ZhaoSWZ22}.
        Closely related to NCL, the self-supervised approach of contrastive learning~\cite{SchroffKP15,ChenK0H20} trains two models to produce latent features that are diverse for false positive cases and similar in true positives through a loss function such as the triplet loss that enforces the models to learn the similarity metric.
    
        \begin{figure}
            \includegraphics[width=\linewidth]{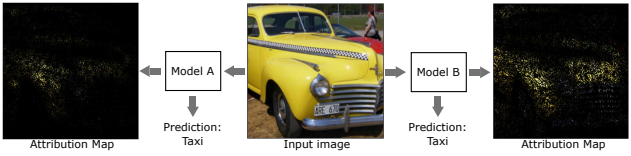}
    	\caption{Attribution map diversity: Two models may predict the same outcome but based on different \textit{evidence}.}\label{fig2}
        \end{figure}
    
        \textbf{Input feature attribution diversity}
        The importance of input features can be used to measure behavior diversity that to the best of our knowledge has not been explored.
        Figure~\ref{figDivTypes} shows the relationship of an attribution-based metric w.r.t. prediction and representation diversity.
        Note, that attribution is not the same as attention or attributes in the context of zero-shot learning.
        Attention maps, such as those obtained from the activation of intermediate layers of CNNs, reflect the excitation of a network given an input.
        This activation however is not necessarily correlated with the final prediction, e.g., it could be an inhibitory factor.
        Attribution, on the other hand, indicates the importance of a feature to the final decision. See Figure~\ref{fig2} where \textit{Saliency} is used to display the original pixels masked by the attribution scores from each model. A change to a pixel with high attribution (brighter) will have a stronger influence on the model prediction than a change to a pixel with low attribution.
    
    \subsection{Other diversity-based resilience approaches}
        Augmenting the training data by applying affine transformations such as rotations and scaling, geometric distortions such as blurring, and texture transfer help DL models to generalize better with a limited training data~\cite{dataAugMiko}.
        Adversarial training increases the robustness to intended attacks with adversarial samples to limit the model vulnerability to input perturbations~\cite{GoodfellowSS14}.
        Such training data approaches are effective and complementary to the design diversity approaches of this study that address the model diversity.
    
        Modality and point of view diversity~\cite{multimodal} is an approach to address the failure modes of sensors such as cameras, radar, and lidar.
        The design diversity of DL models explored in this study is orthogonal to this approach, as model diversity can be applied to every single modality.

\section{Methodology}
\label{labelDesignDiversity}

    We propose three sets of experiments:
        
    \begin{enumerate}
        \item Evaluate resiliency of diverse architectures and training approaches.
        \item Measure diversity of prediction and diversity of attribution from independently created models of diverse architectures and evaluate robustness correlation.
        \item Enforce diversity with NCL, evaluate the resulting robustness, and inspect three kinds of diversity: prediction, representation, and attribution.
    \end{enumerate}	
    
    In addition to addressing the main research questions, we put the following hypothesis to test: that attribution-based diversity (Equation~\ref{eq:attMetric}) can be positively correlated with ensemble resiliency if a better accuracy trade-off is achieved compared to prediction-based diversity.
    \begin{equation}
        A = {\sum_{c=1}^{C}{\sum_{p=1}^{P}{Var(a_{c,p})}}}
        \label{eq:attMetric}
    \end{equation}
    where $a$ is the input attribution score of a model at color channel $c$ and pixel coordinate $p$.
    The computation of the input attribution scores $a$ is performed with an attribution method, such as Saliency. 

    \begin{figure}
	\includegraphics[width=\linewidth]{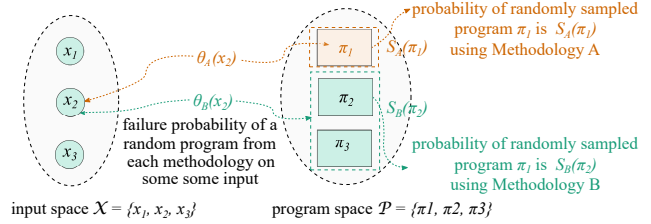}
	\caption{Relation between diverse design methodologies and the difficulty function in the LM Model~\cite{LMmodel}} 
        \label{fig_designdiv_space}
    \end{figure}
        
    This hypothesis is inspired by the theoretical result of the Littlewood and Miller (LM) model~\cite{LMmodel} that diverse design choices can produce less common failures. Diverse attribution maps of correct classifications imply that the models make predictions based on independent factors, which is not the case in prediction diversity.

    \subsection{Probability model for design diversity (LM model)}
            
        The Littlewood and Miller (LM) model~\cite{LMmodel} defines a probabilistic framework to analyze the impact of methodological diversity in the expected failure behavior.
        The model defines: 1) an input space $\mathcal{X}=\{x_1,x_2,...\}$, representing all possible inputs $x$ to a program and 2) a program space $\mathcal{P}=\{(\pi_1, \pi_2)\}$, for all possible programs $\pi$ that could implement a program specification.
        A given design methodology will determine the probability to come up with a program $\pi$ and is denoted as $S_{A}(\pi)$.
        Another design methodology $S_{B}(\pi)$ will assign a different probability to the same program.
        The model uses the concept of a difficulty function  $\theta_{M}(x)$ that measures the probability that a randomly chosen program $\pi$ from a given methodology distribution $S_{M}(\pi)$ will fail on a particular input $x \in \mathcal{X}$.
        The key insight consists in noticing that $\theta_{A}(x)$ can be different for a different methodology $\theta_{B}(x)$, i.e., for some methodology, a certain input may be difficult, but for another, it may be easy.
        See Figure~\ref{fig_designdiv_space} for a visual representation of these spaces.
        An analysis of this model concludes that if the design methodologies produce different difficulty functions $\theta$, then the expected failure behavior on a random input will be negatively correlated due to the fact that the covariance of the $\theta$'s can be negative.
            
        With this model, it is finally shown that a design methodology with diverse design choices that satisfy the following three properties will result in less common failures: 1) logically unrelated (one decision is independent of the other), 2) common failures of a decision are due to different factors, and 3) indifference to the selection of each methodology (no methodology is superior).

    \subsection{Loss function to enforce attribution diversity}
	\label{labelLossFunction}
    
        We perform a first attempt to enforce attribution diversity with the following loss:
        \begin{equation}
            \frac{1}{M}\sum_{i=1}^{M}{l(h^{i},y)}-\lambda{A}
            \label{eq:attDivEnf}
        \end{equation}
    
        This loss computes attribution scores variance in an ensemble and uses it as a penalty term weighted by $\lambda$.

    \subsection{Failures addressed}
    
        In this study, we evaluate resilience to covariate dataset distribution shifts, i.e., when the distribution of input features of the test dataset does not match the distribution of the training dataset. 
        We use four natural image perturbations from the \imagenetcbar dataset~\cite{MintunKX21}   that are sensible to occur in vision application domains, such as obstructions or liquid contaminants.
        Our scope is not to evaluate robustness against adversarial attacks, label shift variations, or resiliency to noise variations such as Gaussian, brown, etc.

\section{Experimental results}
\label{labelExperiments}

    \subsection{ML resiliency to data corruptions}
    \label{labelExpMlresiliency}

        To understand the relationship between accuracy, size, and resiliency to natural corruptions of DL models we evaluate a set of architectures (convolutional NNs, transformers, and subnetworks from neural architecture search (NAS)) and training approaches (supervised, self-supervised, and knowledge distillation) on both the ImageNet validation dataset and on the corrupted version \imagenetcbar "Lines" (strength of 1.6). See Table~\ref{tab:archCompa}.

        \begin{table}
            \footnotesize
            \caption{Resiliency of architectures to natural image corruption (\imagenetcbar Lines(1.6 strength))}\label{tab:archCompa}
            \resizebox{\linewidth}{!}{%
            \begin{tabular}{|c|c|c|c|c|c|c|c|c|}
            \hline
            Training & Arch. & DL model & \#  & \multicolumn{2}{c|}{ImageNet}  & \multicolumn{2}{c|}{\imagenetcbar}  \\
            approach & class &  & Params  &  \multicolumn{2}{c|}{} &  \multicolumn{2}{c|}{Lines1.6} \\
             &  & & & top1 & top5 & top1 & top5\\
            \hline
            Self-sup. & CNN &  DINO ResNet50~\cite{CaronTMJMBJ21} & 25.6M & 75.30 & 92.61 & 21.80 & 40.66\\
            Superv. & CNN &  ResNet50~\cite{HeZRS16} & 25.6M & 75.85 & 92.88 & 34.95 & 55.88\\
            Superv. & CNN &  ResNext50\_32\_4d~\cite{XieGDTH17} & 25.0M & 77.49 & 93.57 & 39.46 & 60.61\\
            Superv. & CNN &  ResNet152~\cite{HeZRS16} & 60.2M & 78.25 & 93.96 & 40.17 & 62.09\\
            Superv. & NAS &  EfficientNet\_b7~\cite{TanL19} & 66.3M & 74.82 & 92.13 & 51.11 & 73.81\\
            Superv. & CNN &  ResNext101\_64x4d~\cite{XieGDTH17} & 83.4M & 82.90 & 96.22 & 51.89 & 72.46\\
            Superv. & ViT &  Swin tiny~\cite{LiuL00W0LG21} & 28.3M & 81.34 & 95.612 & 55.29 & 78.45 \\
            Self-sup. & ViT &  DINO ViT\_b\_8~\cite{CaronTMJMBJ21} & 87.3M & 80.06 & 95.02 & 55.75 & 78.37\\
            Superv. & ViT &  Swin base~\cite{LiuL00W0LG21} & 87.8M & 83.30 & 96.46 & 58.71 & 81.05\\
            Superv. & ViT &  ViT base p16~\cite{DosovitskiyB0WZ21} & 86.6M & 80.88 & 95.28 & 61.28 & 82.26\\
            K. distill. & ViT &  DeiT-base~\cite{TouvronCDMSJ21} & 87.3M & 83.34 & 96.50 & 64.07 & 84.66\\
            Self-sup. & ViT &  SwinV2~\cite{Liu0LYXWN000WG22} & 87.9M & 83.34 & 96.44 & 59.976 & 81.70\\
            \hline
            \end{tabular}
        }
        \end{table}
        
        \textbf{Observations to Table~\ref{tab:archCompa}:} Although the model size is highly correlated with the final accuracy and resilience in the corrupted dataset, the architecture seems to play a more determining factor.
        The smallest transformer with only 28M parameters is superior to other CNNs with 2 or 3x more parameters.
        Self-supervision slightly decreases both metrics as appreciated in the comparison of ResNet50 and ViT models using supervised learning.
        SwinV2 is an exception but this model introduced more architectural innovations too.
        Knowledge distillation from a CNN teacher shows a slight improvement over supervised ViTs.

\begin{figure}
            \includegraphics[width=\linewidth] {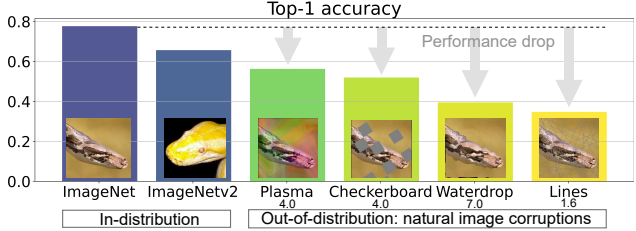}
		\caption{Top-1 accuracy of ResNet50 on in-distribution data sets (ImageNet and ImageNetv2\cite{RechtRSS19}) and out-of-distribution datasets (\imagenetcbar\kern-0.5em). The resilience of ResNet50 drops significantly against natural corruptions.} \label{fig:indivAcc}
	\end{figure}

        To understand the effect of other corruptions, we evaluate a ResNet50 model\footnote{In the remainder of this paper, we select ResNet50 as the baseline architecture due to its common use as a reference.} on six different data sets: ImageNet~\cite{DengDSLL009}, ImageNetv2~\cite{RechtRSS19}, and four corruptions on a fixed perturbation strength from the \imagenetcbar dataset~\cite{MintunKX21}: Plasma (4.0), Checkerboard(4.0), Waterdrop(7.0) and Lines (1.6). See Figure~\ref{fig:indivAcc}.        
        The first two are in-distribution, i.e., the covariates (input features) and labels of the validation set follow a similar distribution to the training data set.
        The last four are out-of-distribution, as the model has never seen such corruptions of the input images during training.
            
        \textbf{Observations to Figure~\ref{fig:indivAcc}:} Different corruptions have different effects on the model performance, and a typical ``good" classifier with accuracy close to 80\% can have a tremendous performance decrease in situations where a human would probably not.
		
    \subsection{Diversity of ensembles from heterogeneous architectures}
    \label{labelExpHetero}

        To understand the diversity/accuracy trade-off of the attribution-based metric in comparison to the established prediction-based diversity approach we perform two different experiments: First, we create multiple ensembles of independently trained models with a wide diversity in architecture. Second, we create multiple ensembles using models discovered in a weight-sharing super-network~\cite{CaiGWZH20}, i.e., models whose architecture has been found using neural architecture search (NAS) and not by manual design.

        The architectures explored here are CNNs (ResNext~\cite{XieGDTH17} \& SqueezeNet~\cite{IandolaMAHDK16}), Vision transformers (DeiT~\cite{TouvronCDMSJ21}) and NAS (MNASNET~\cite{TanCPVSHL19} \& BootstrapNAS~\cite{munoz2022enabling}) using supervised or self-supervised training\footnote{Details on the architecture and optimization hyper-parameters can be found in Section A of the supplementary material.}. In total 14 models were trained with different hyperparameters to create three-member ensembles of all possible combinations.

\begin{figure}[htp]
            \subcaptionbox{Attribution diversity\label{fig:hLa}}{\includegraphics[width=0.5\linewidth]{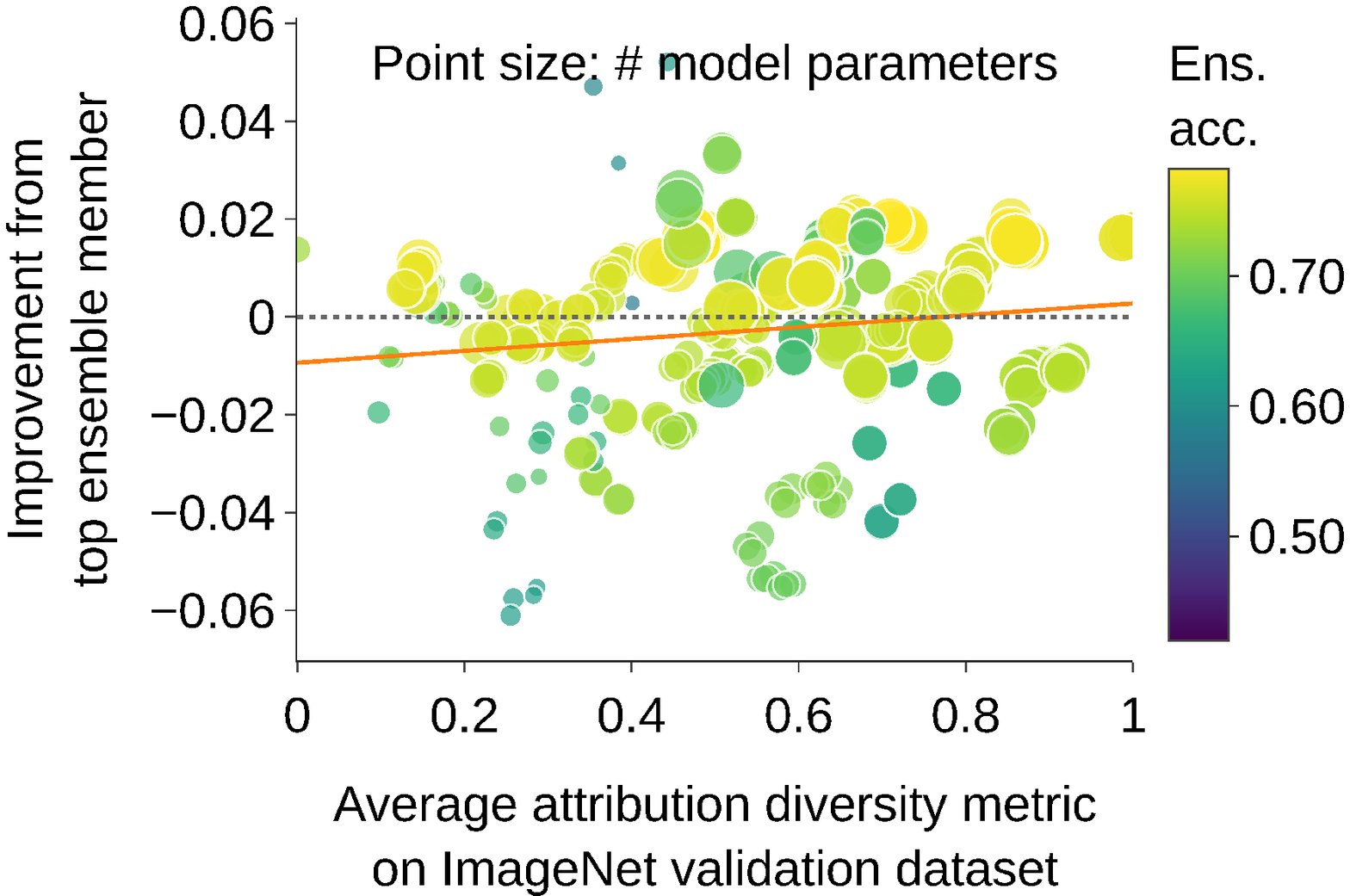}}\hspace{0em}%
            \subcaptionbox{Prediction disagreement\label{fig:hLb}}{\includegraphics[width=0.5\linewidth]{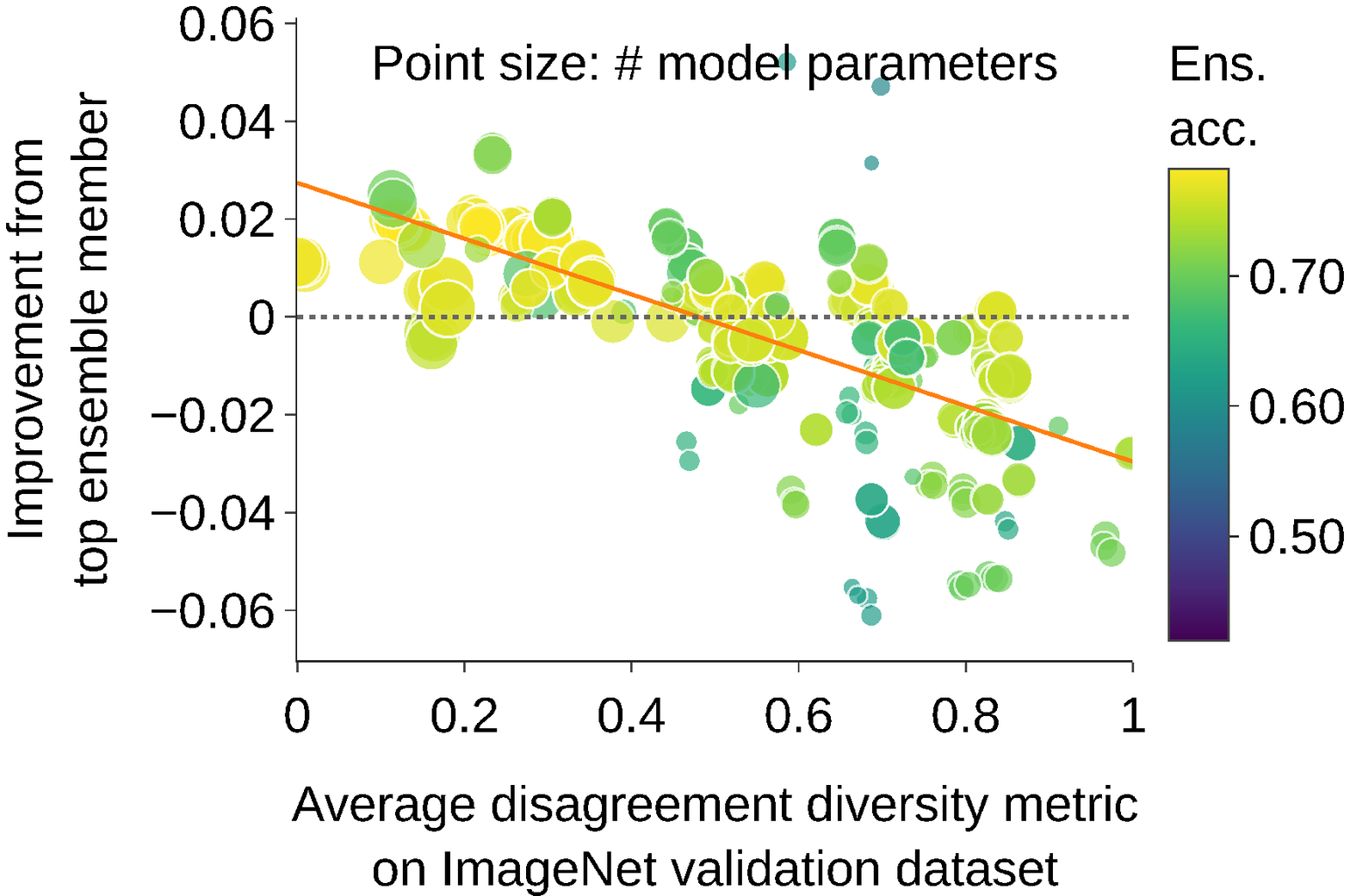}}
            \caption{Evaluation on ImageNet validation dataset of 364 three-member ensembles from heterogeneous architectures using \textbf{averaging as consensus mechanism}. Y-axis: Improvement of the ensemble against its own top ensemble member. X-axis: Normalized diversity metric. Color: absolute ensemble accuracy. Bubble size: Model parameter size. The attribution diversity metric is not negatively correlated with the ensemble improvement as disagreement diversity is.}
        \label{fig:heteroDisLinear}
        \end{figure}

        Figure~\ref{fig:heteroDisLinear} shows the ensemble performance of all 364 ensembles created from these 14 models using an averaging consensus mechanism, i.e., the logit output of all ensemble members is averaged first and then the highest score is used to make the prediction.
        The left-hand side shows on the X-axis the attribution-based diversity metric (Eq.~\ref{eq:attMetric}).
        The right-hand side shows the disagreement prediction diversity metric (Eq.~\ref{eq:disagreement}).
        Each point is an ensemble evaluated on the entire validation dataset of ImageNet.
        The color indicates the final average accuracy of the ensemble.
        The Y-axis indicates the average benefit of creating an ensemble: $Y= A_{ens}-A_{top}$, where $A_{ens}$ is the ensemble accuracy and $A_{top}$ is the accuracy of its most accurate member, i.e., how much accuracy improvement was obtained in comparison to a single model (the most accurate in the ensemble).
        In this way, it can be appreciated when an ensemble makes sense: it has to lay above the zero line (dashed).
        The ensemble cost is measured by the number of parameters which has a direct influence on the memory and the number of operations required.
        The ideal ensemble is one with the brightest color, smallest radius, and residing above zero.
        
        \textbf{Observations to Figure~\ref{fig:heteroDisLinear}:} The figure shows how attribution diversity is positively correlated with ensemble improvement, while it corroborates the known fact that prediction-based diversity is negatively correlated~\cite{KunchevaW03}.

\begin{figure}[htp]
            \subcaptionbox{Attribution diversity\label{fig:hDVa}}{\includegraphics[width=0.5\linewidth]{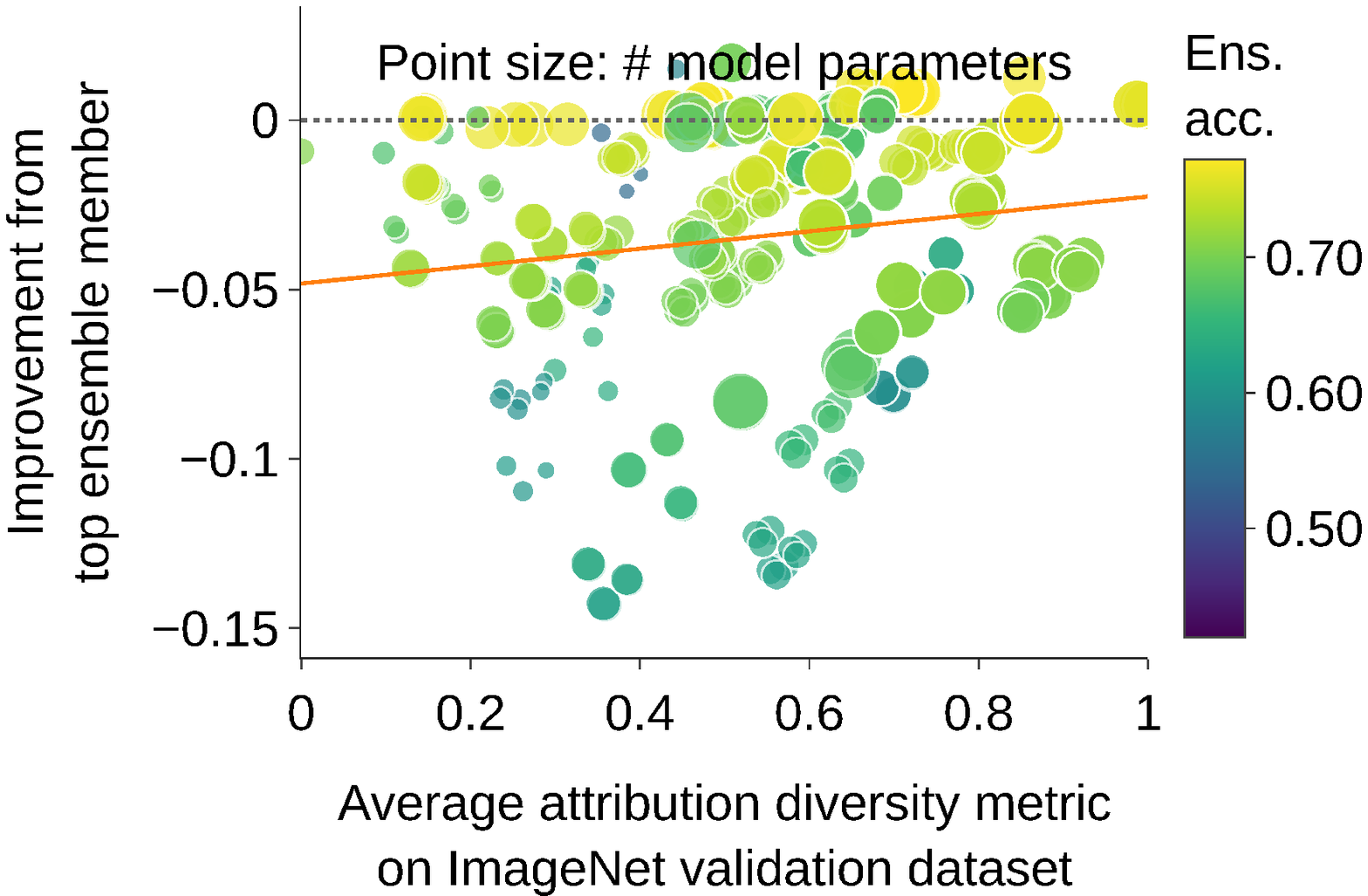}}\hspace{0em}%
            \subcaptionbox{Prediction disagreement\label{fig:hDVb}}{\includegraphics[width=0.5\linewidth]{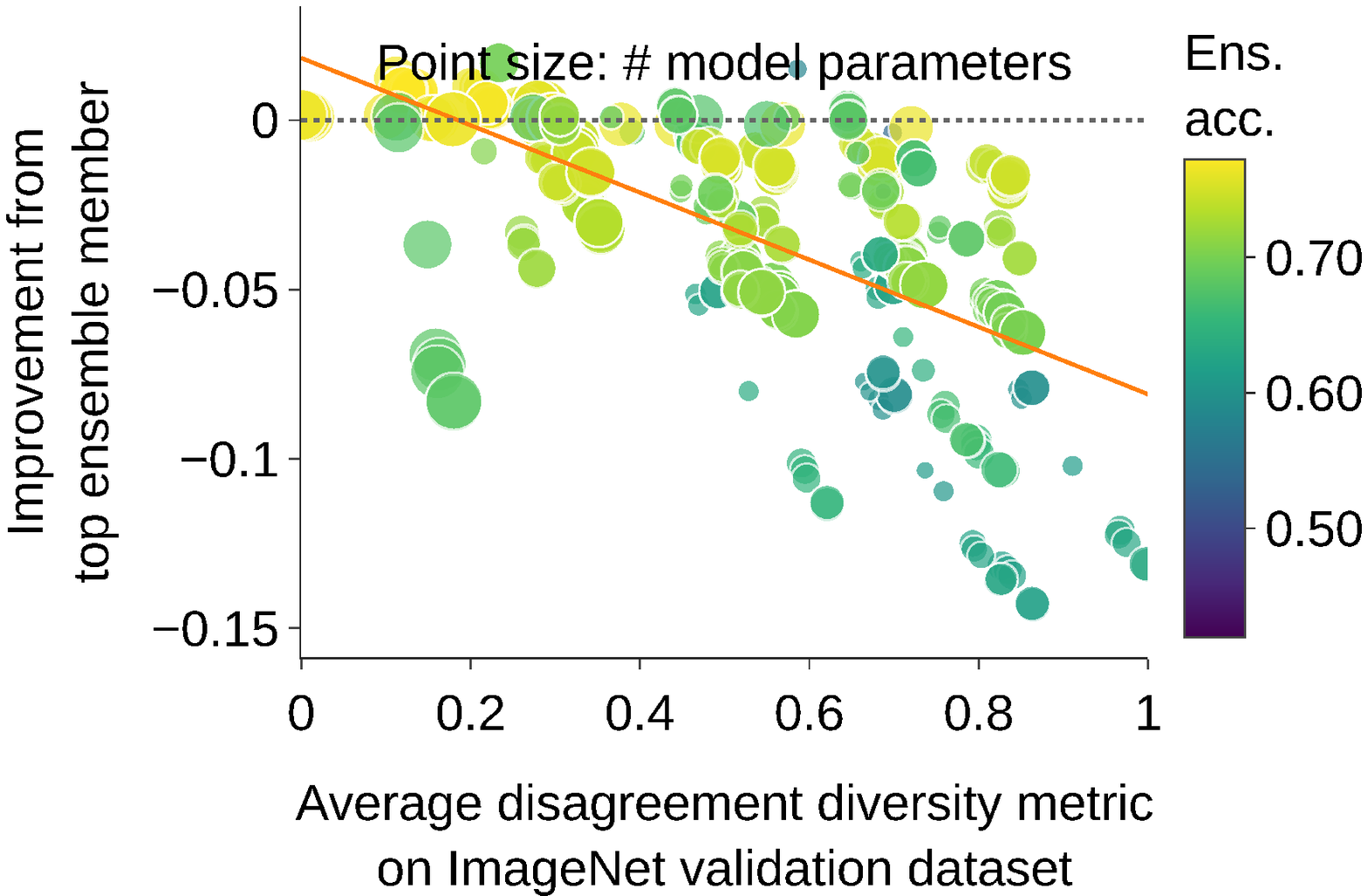}}
            \caption{Evaluation on ImageNet validation dataset of the same ensembles as in Figure~\ref{fig:heteroDisLinear} but using \textbf{voting as consensus mechanism}. In contrast to averaging, voting produces mostly ensembles that decrease the final performance instead of improving it.}
            \label{fig:heteroDisVoting}
        \end{figure}

        Figure~\ref{fig:heteroDisVoting} shows the same ensemble combinations, but this time using a majority voting consensus mechanism, i.e., the prediction with the highest number of votes wins. Draws are randomly resolved.

        \textbf{Observations to Figure~\ref{fig:heteroDisVoting}:} The same correlation trends can be observed with majority voting. However, the most interesting aspect is that the vast majority of the ensembles here reside under the zero line.
        This means that majority voting with three ensembles tends on average to produce less accurate models. This corroborates the findings of~\cite{KunchevaW03}.

\begin{figure}[htp]
            \subcaptionbox{\centering Attr. ImageNetv2\label{fig3:odda}}{\includegraphics[width=0.2\linewidth]{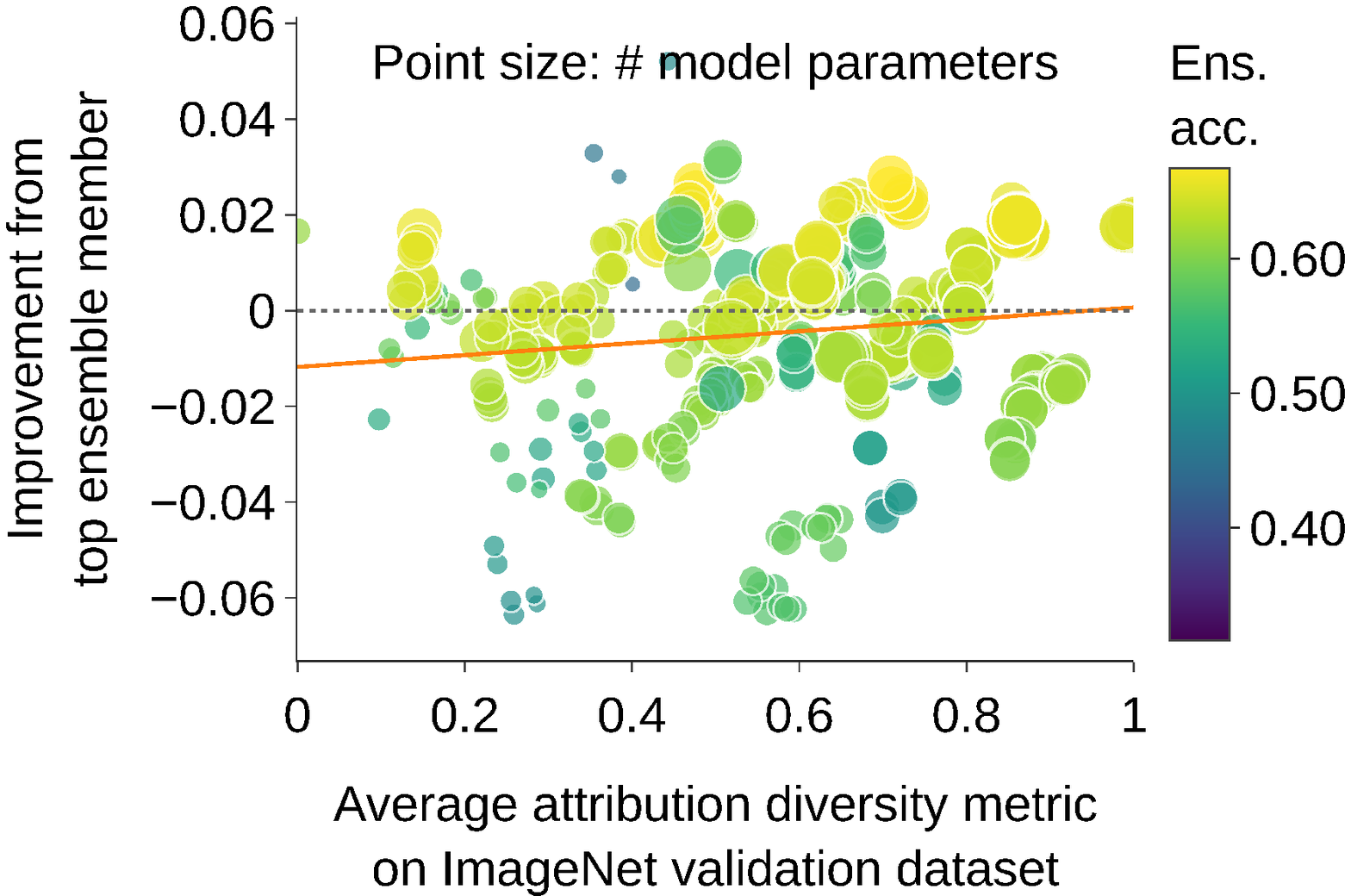}}%
            \hspace{0em}%
            \subcaptionbox{\centering Attr. Waterdrop\label{fig3:oddb}}{\includegraphics[width=0.2\linewidth]{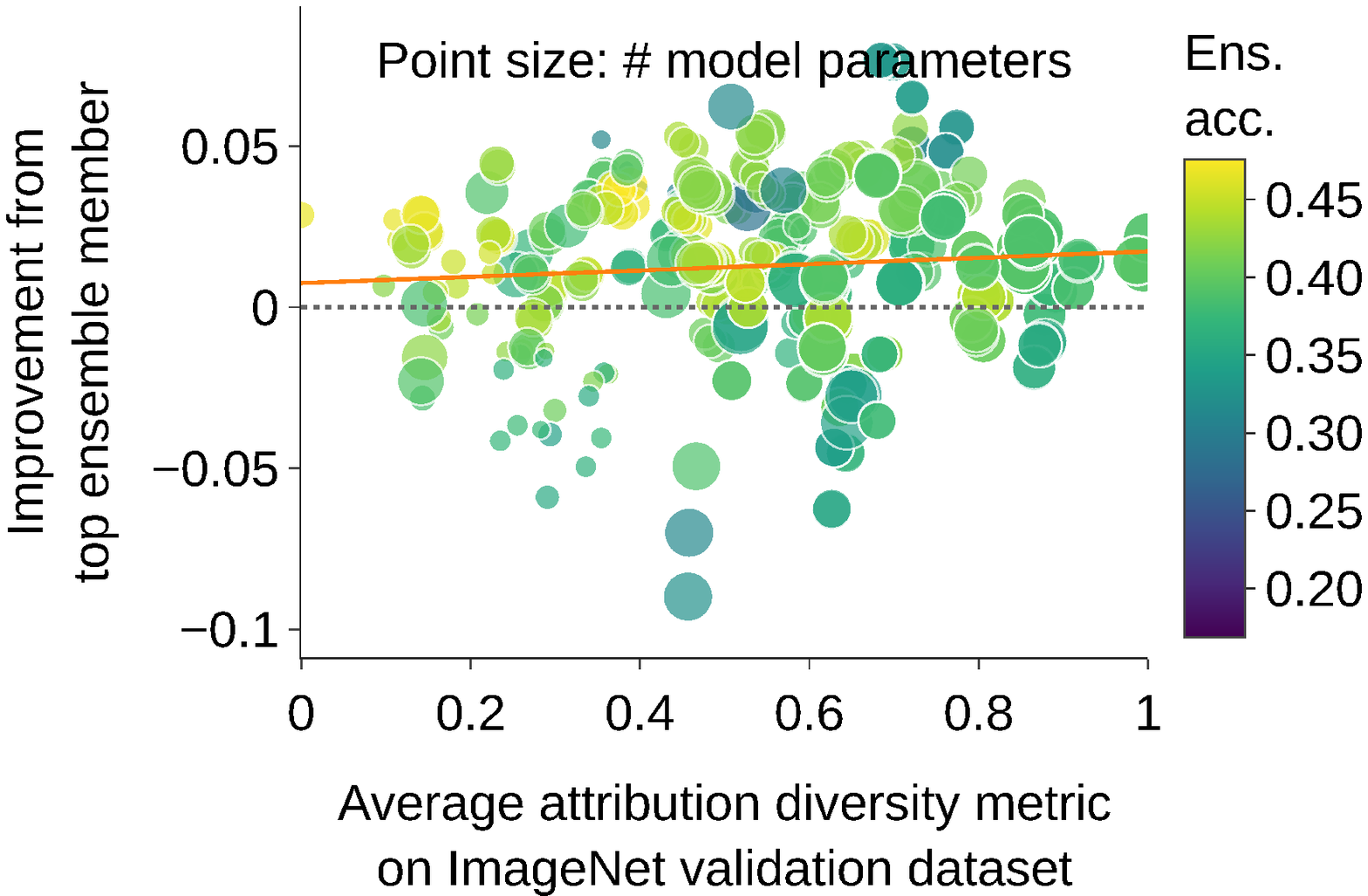}}%
            \hspace{0em}%
            \subcaptionbox{\centering Attr. Lines\label{fig3:oddc}}{\includegraphics[width=0.2\linewidth]{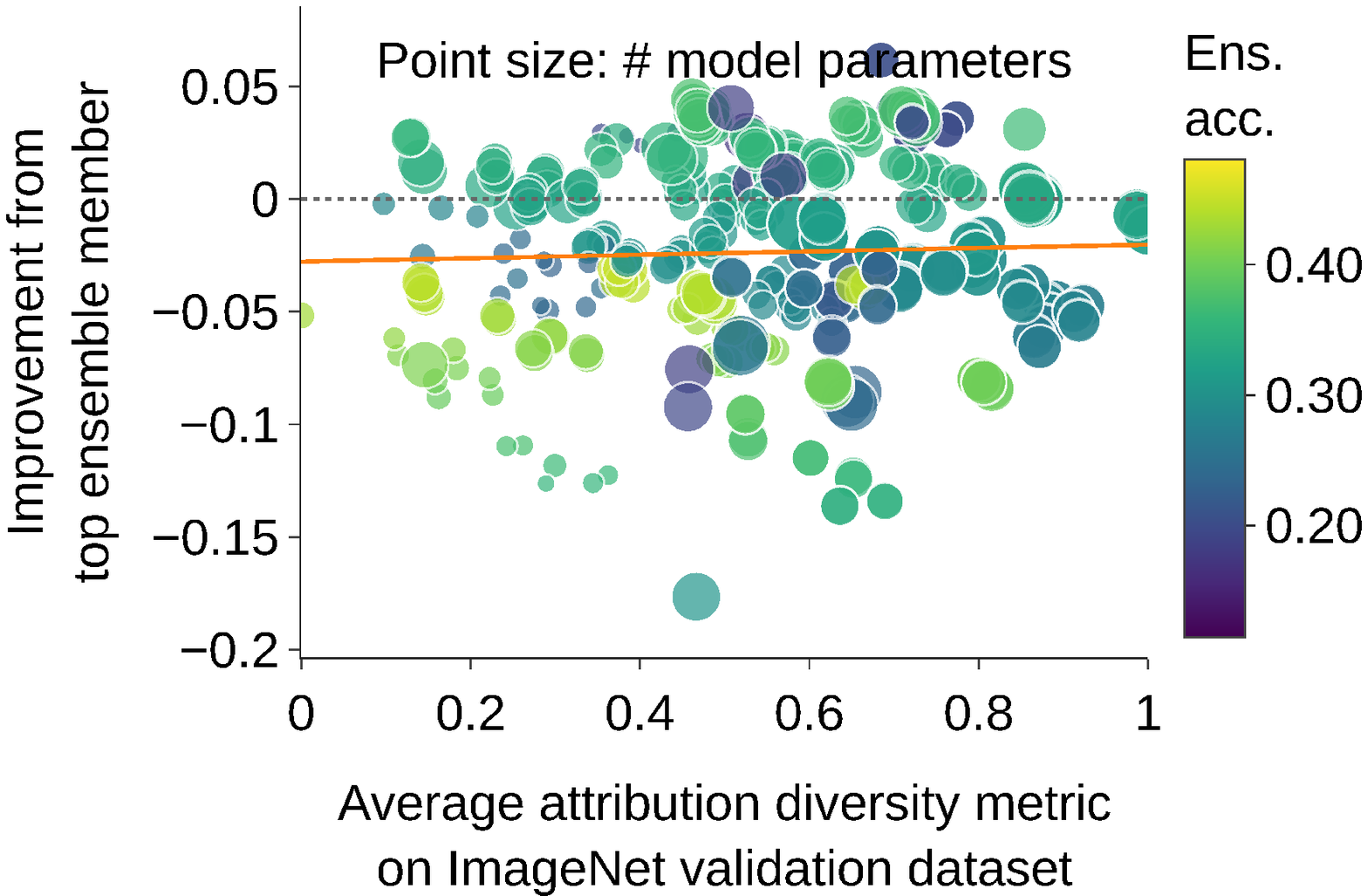}}%
            \hspace{0em}%
            \subcaptionbox{\centering Attr. Plasma\label{fig3:oddd}}{\includegraphics[width=0.2\linewidth]{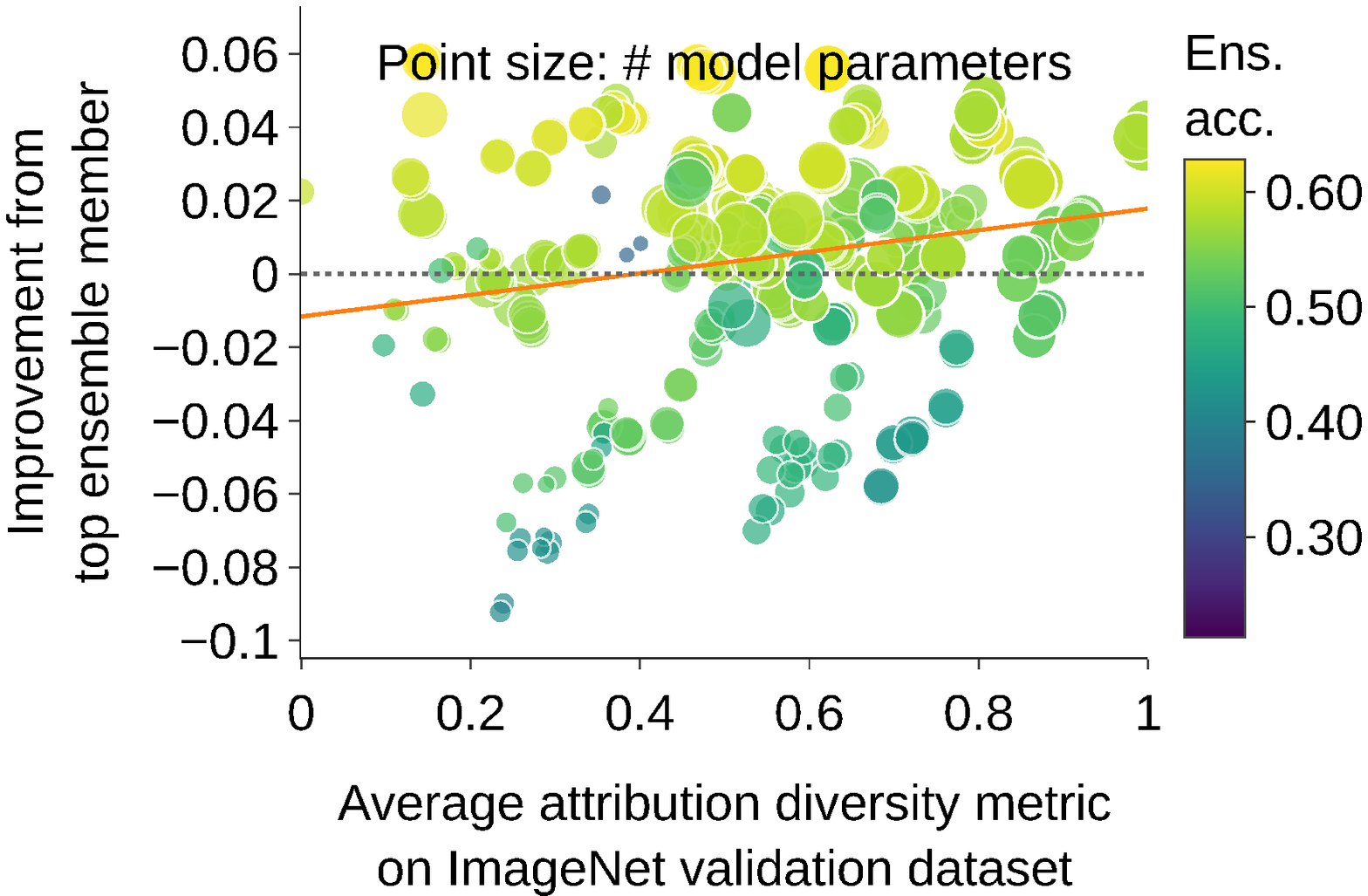}}%
            \hspace{0em}%
            \subcaptionbox{\centering Attr. Checkerb.\label{fig3:odde}}{\includegraphics[width=0.2\linewidth]{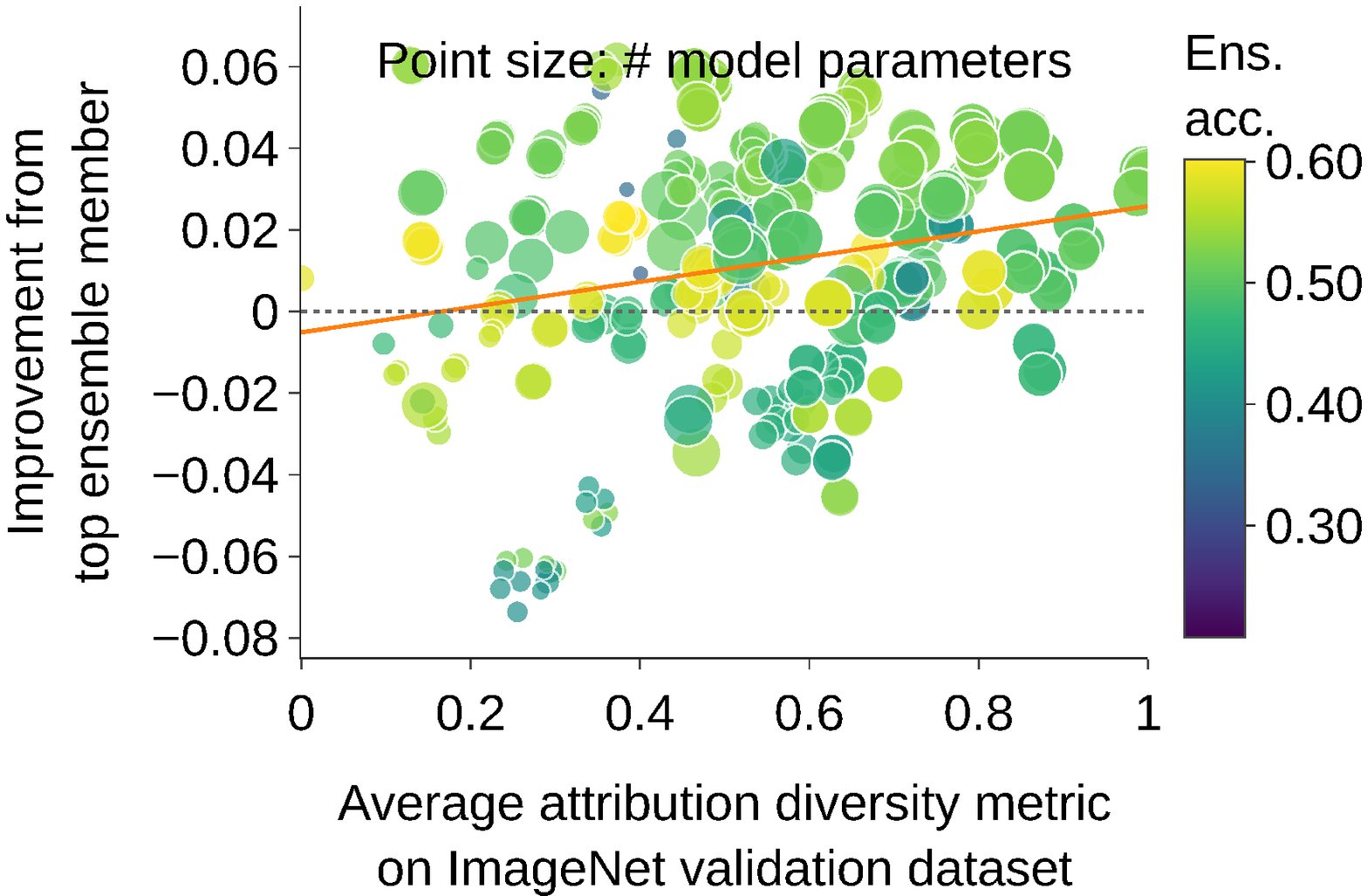}}%
            \hspace{0em}%
            \subcaptionbox{\centering Disagr. ImageNetv2\label{fig3:oddf}}{\includegraphics[width=0.2\linewidth]{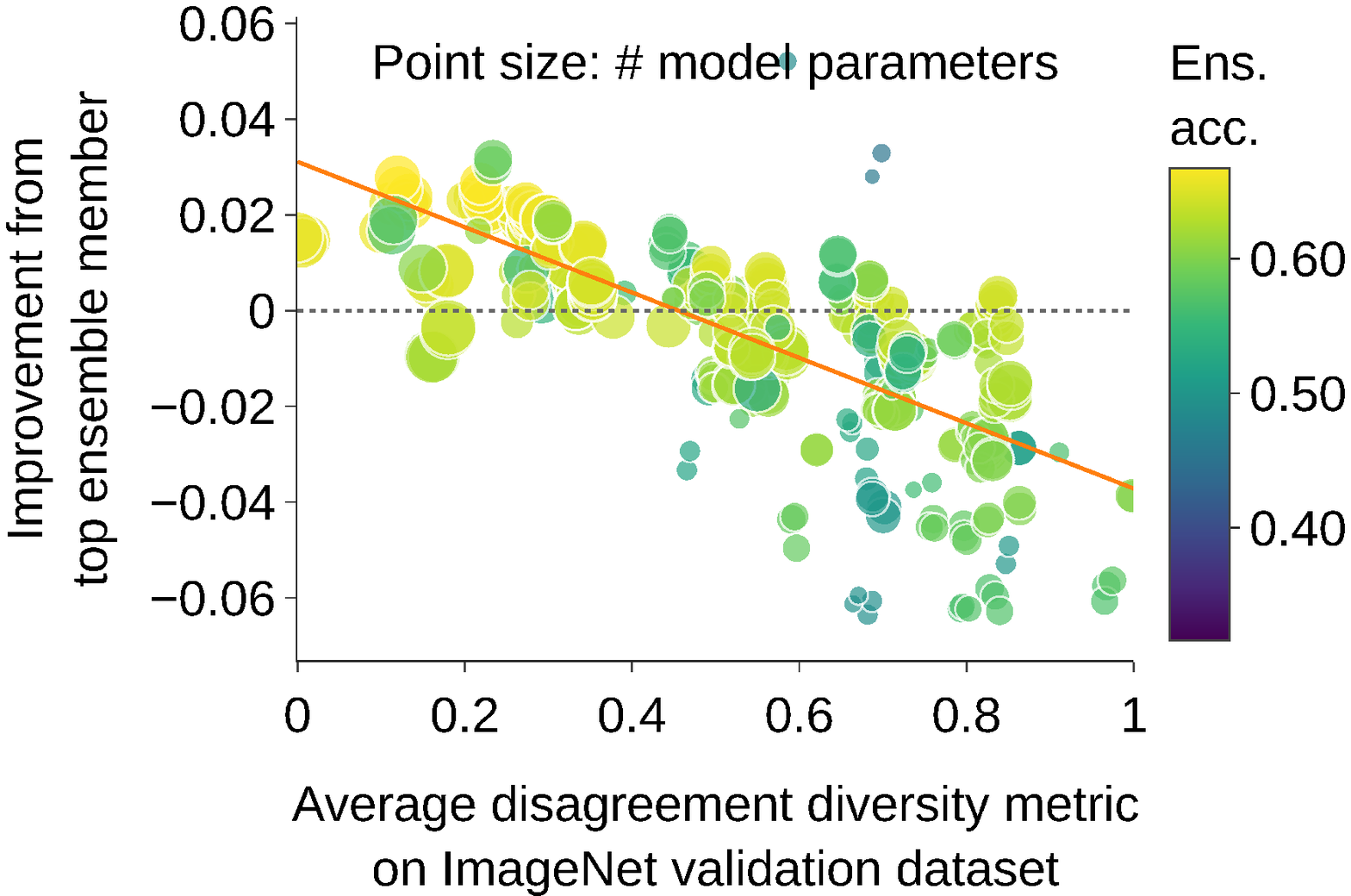}}%
            \hspace{0em}%
            \subcaptionbox{\centering Disagr. Waterdrop\label{fig3:oddg}}{\includegraphics[width=0.2\linewidth]{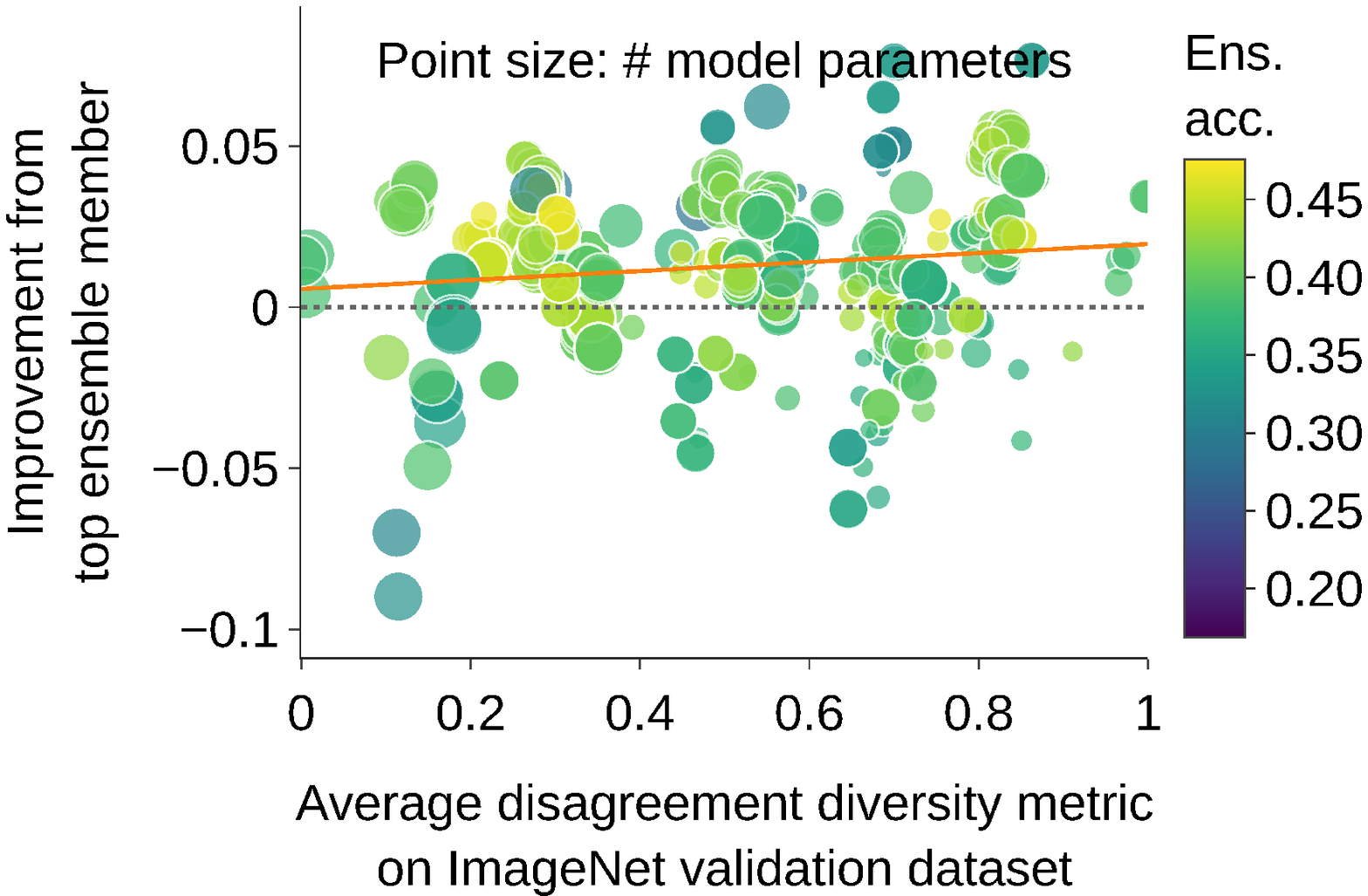}}%
            \hspace{0em}%
            \subcaptionbox{\centering Disagr. Lines\label{fig3:oddh}}{\includegraphics[width=0.2\linewidth]{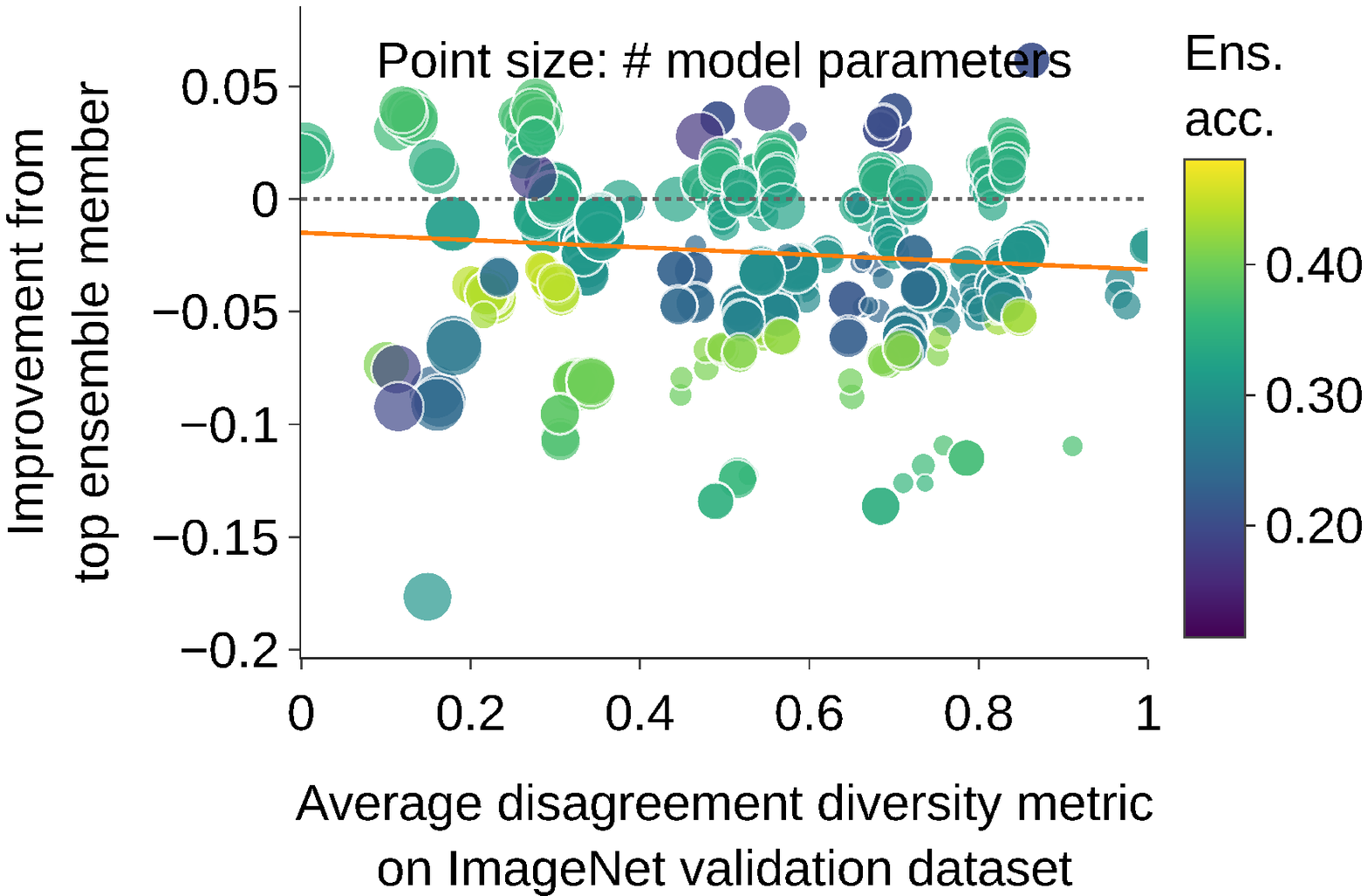}}%
            \hspace{0em}%
            \subcaptionbox{\centering Disagr. Plasma\label{fig3:oddi}}{\includegraphics[width=0.2\linewidth]{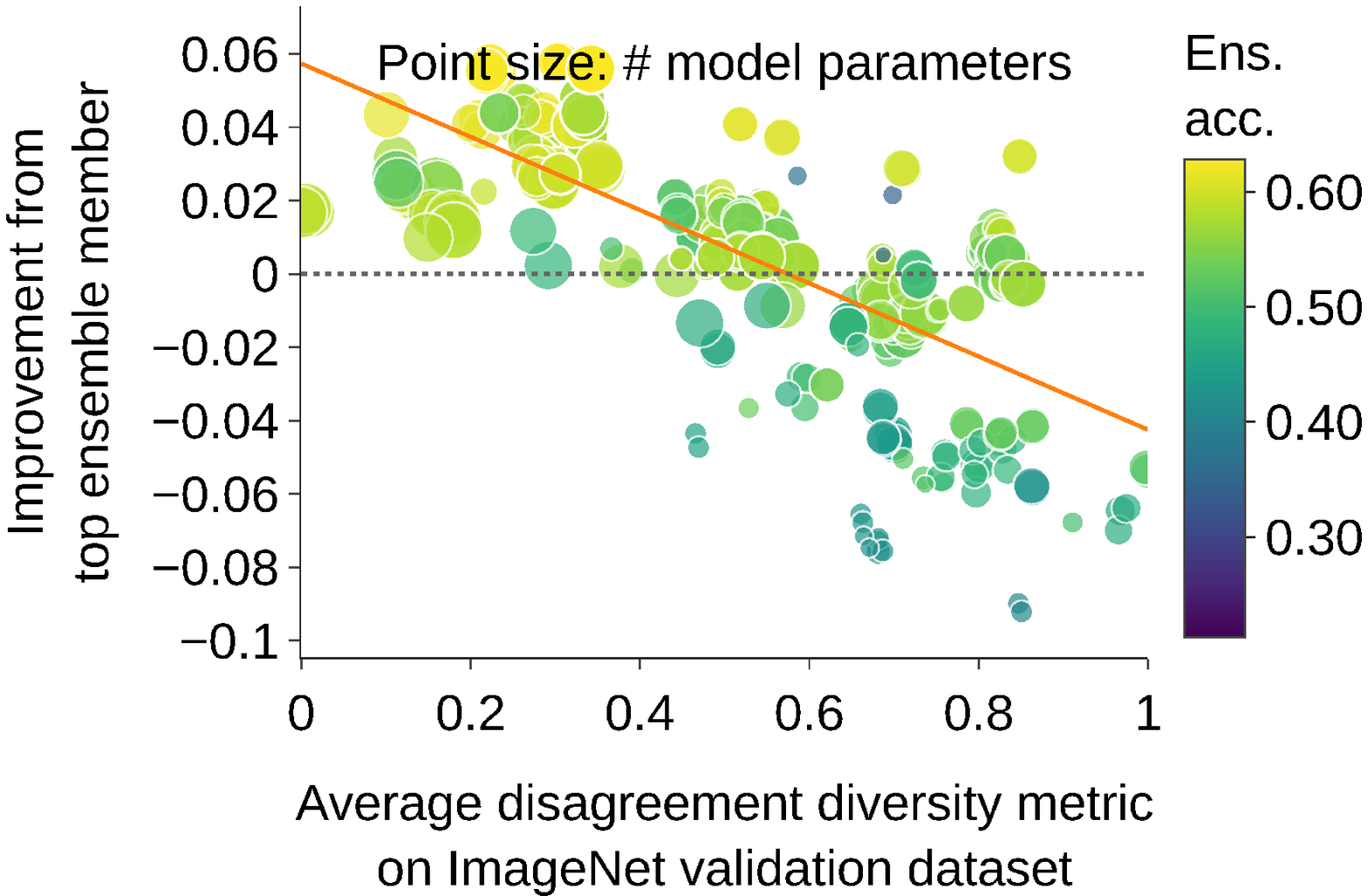}}%
            \hspace{0em}%
            \subcaptionbox{\centering Disagr. Checkerb.\label{fig3:oddj}}{\includegraphics[width=0.2\linewidth]{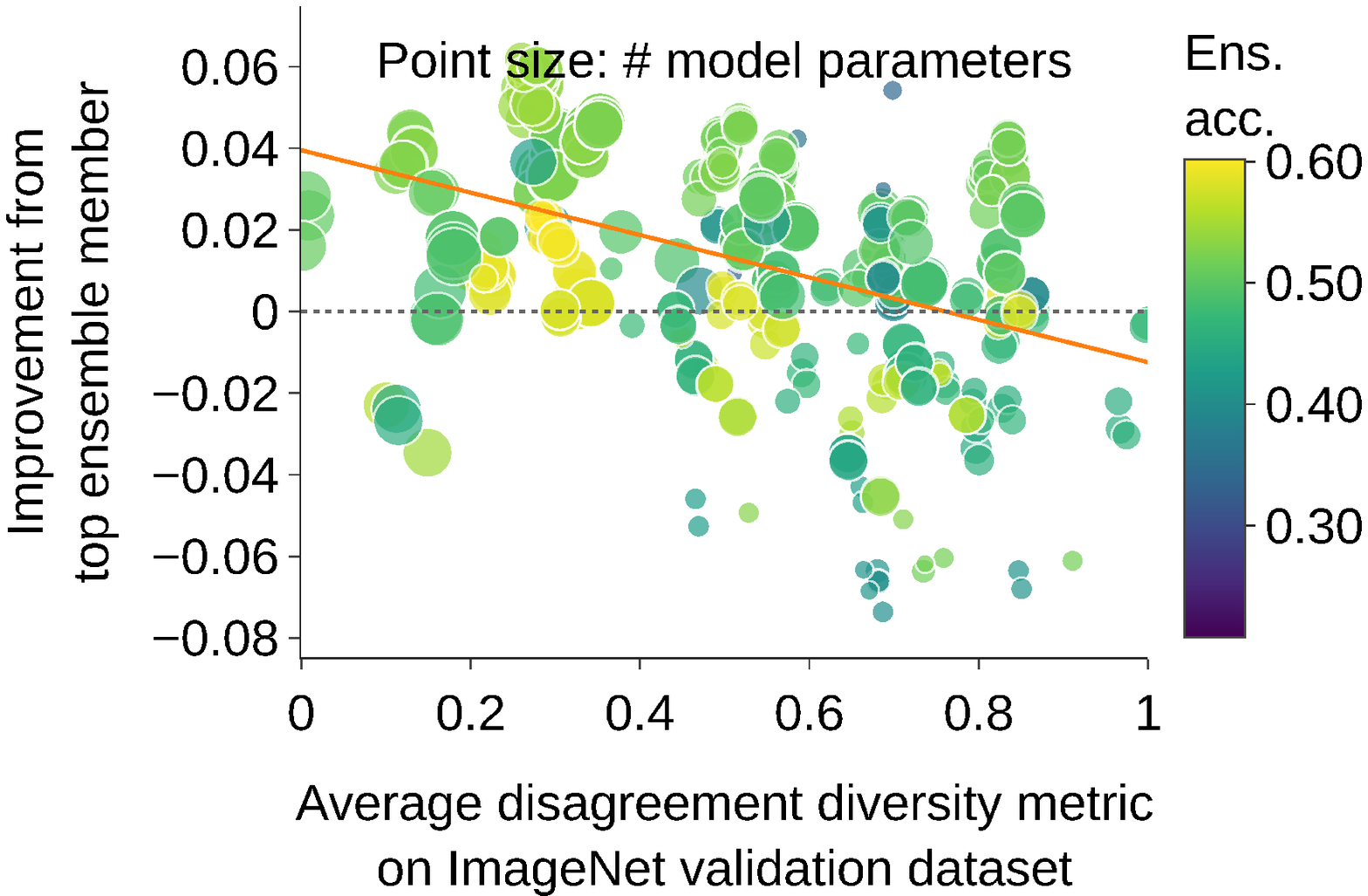}}%
            \hspace{0em}%
            \subcaptionbox{\centering Acc. ImageNetv2\label{fig3:oddk}}{\includegraphics[width=0.2\linewidth]{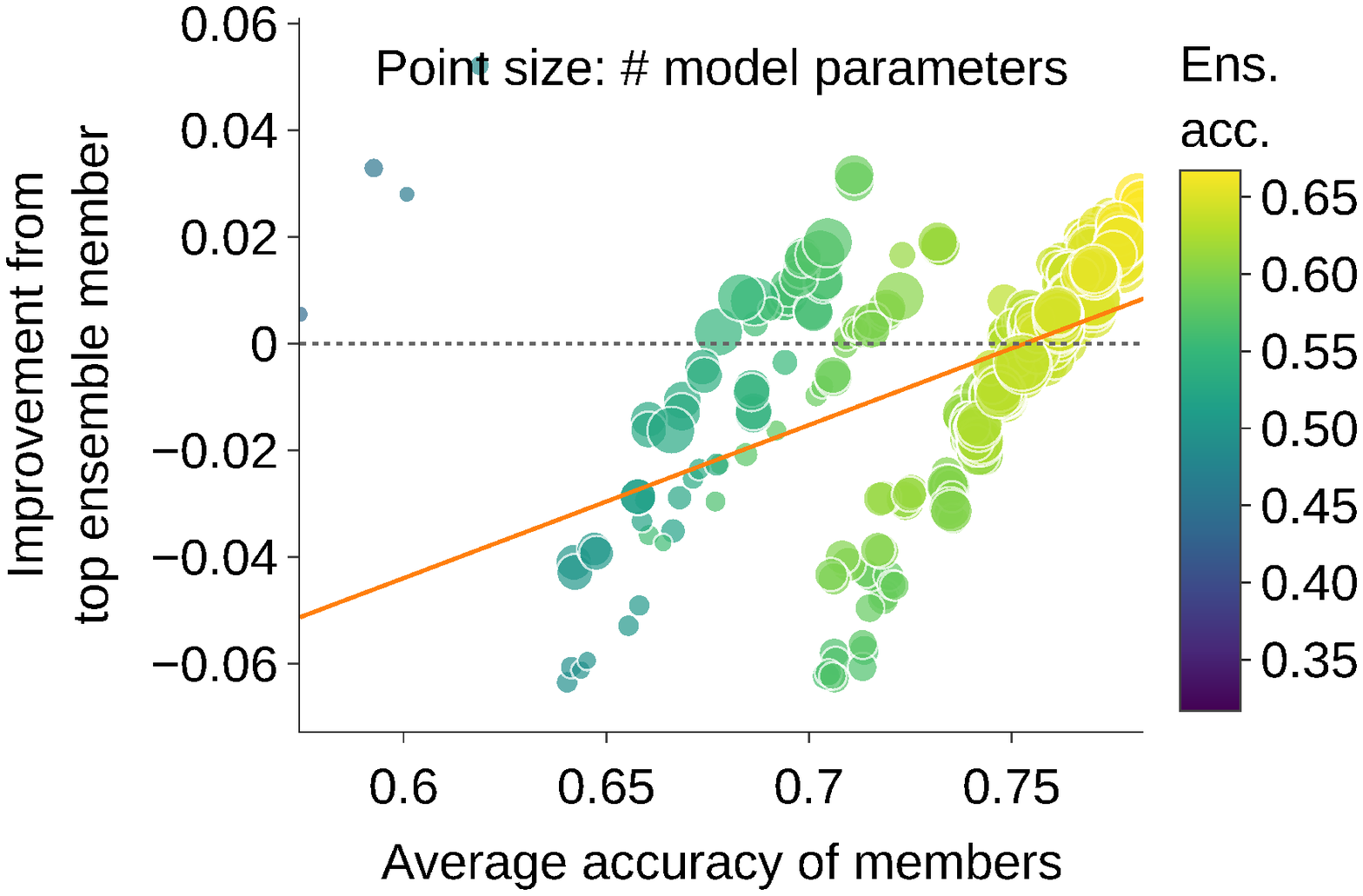}}%
            \hspace{0em}%
            \subcaptionbox{\centering Acc. Waterdrop\label{fig3:oddl}}{\includegraphics[width=0.2\linewidth]{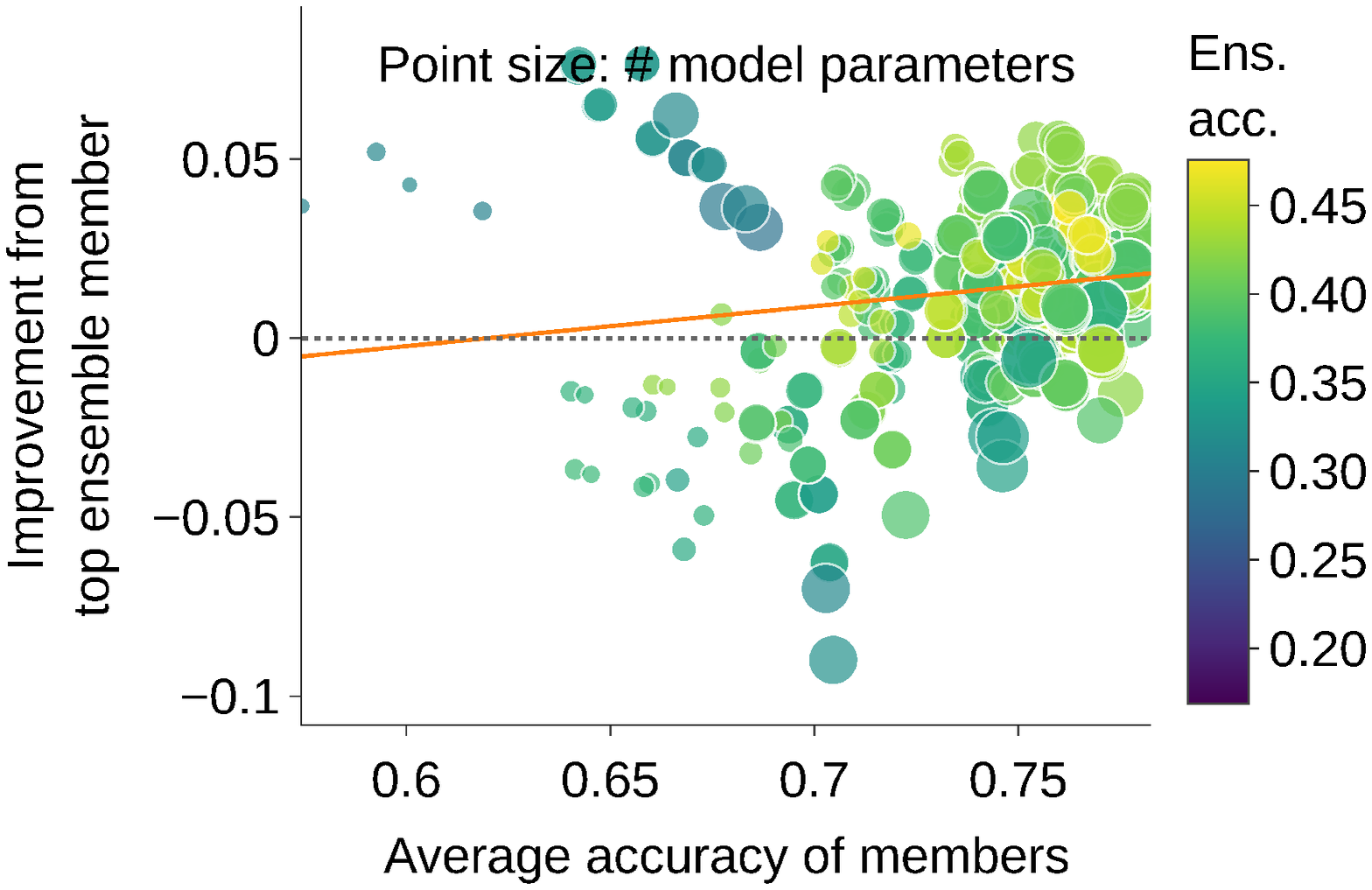}}%
            \hspace{0em}%
            \subcaptionbox{\centering Acc. Lines\label{fig3:oddm}}{\includegraphics[width=0.2\linewidth]{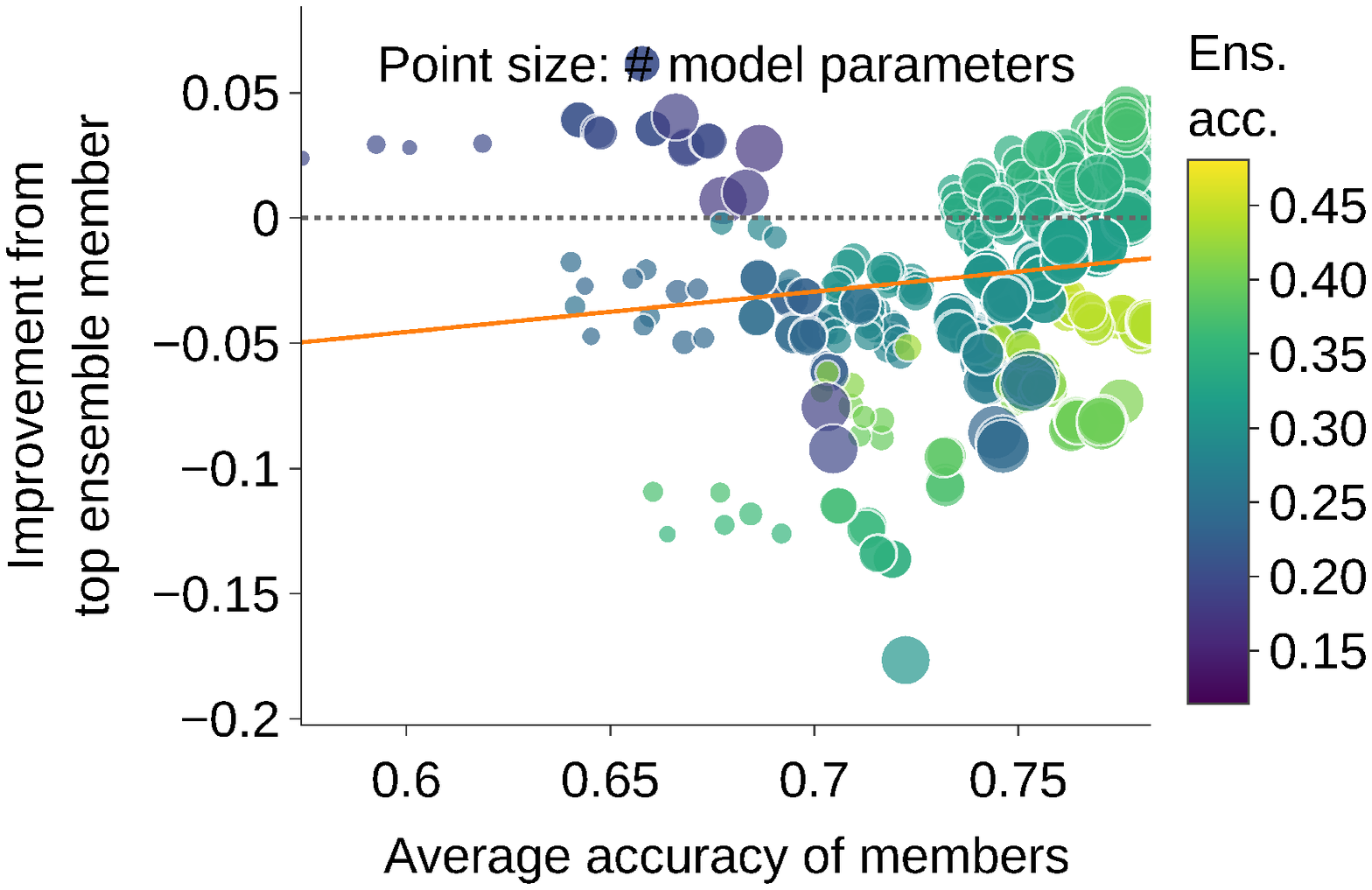}}%
            \hspace{0em}%
            \subcaptionbox{\centering Acc. Plasma\label{fig3:oddn}}{\includegraphics[width=0.2\linewidth]{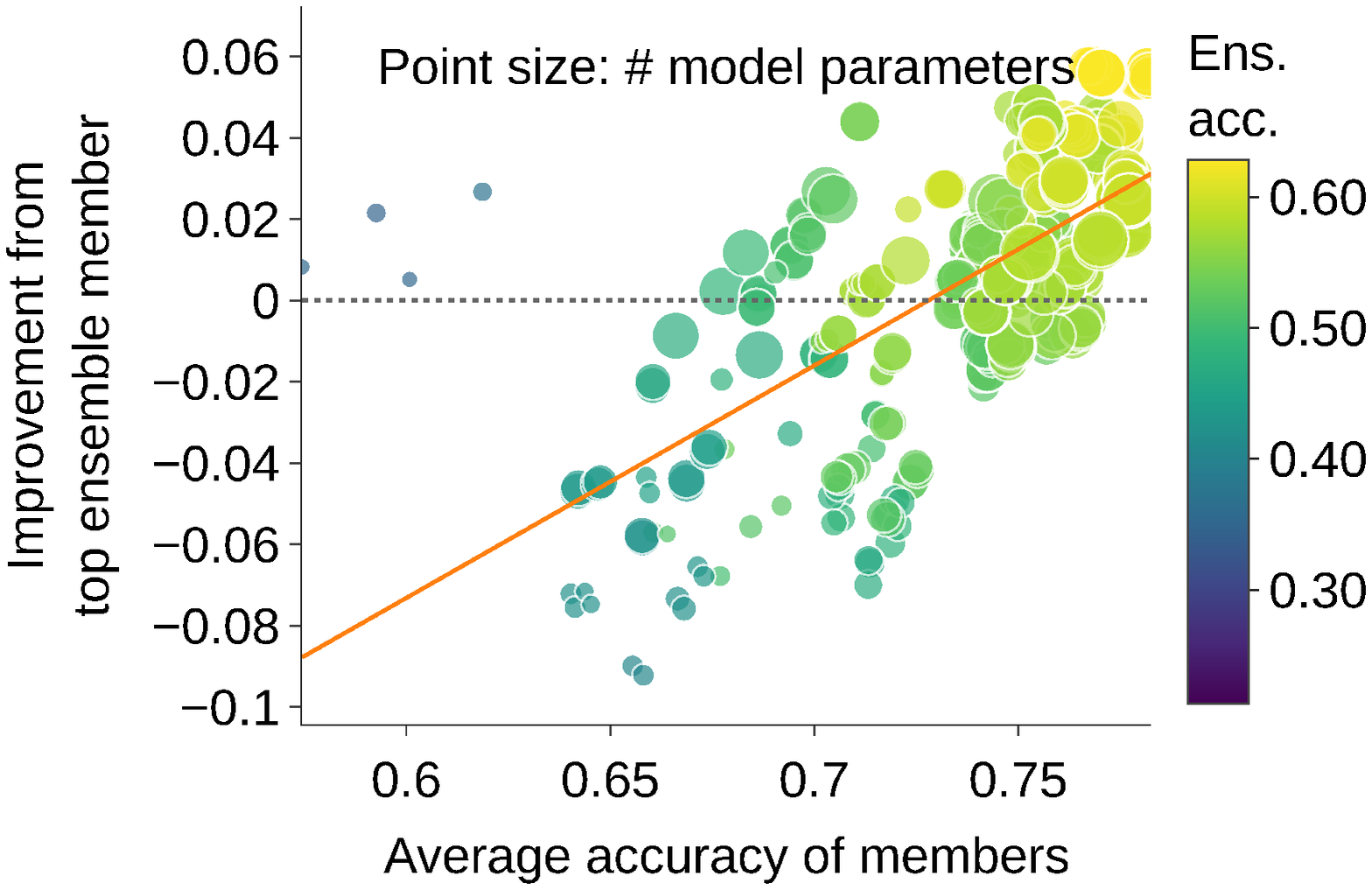}}%
            \hspace{0em}%
            \subcaptionbox{\centering Acc. Checkerb.\label{fig3:oddo}}{\includegraphics[width=0.2\linewidth]{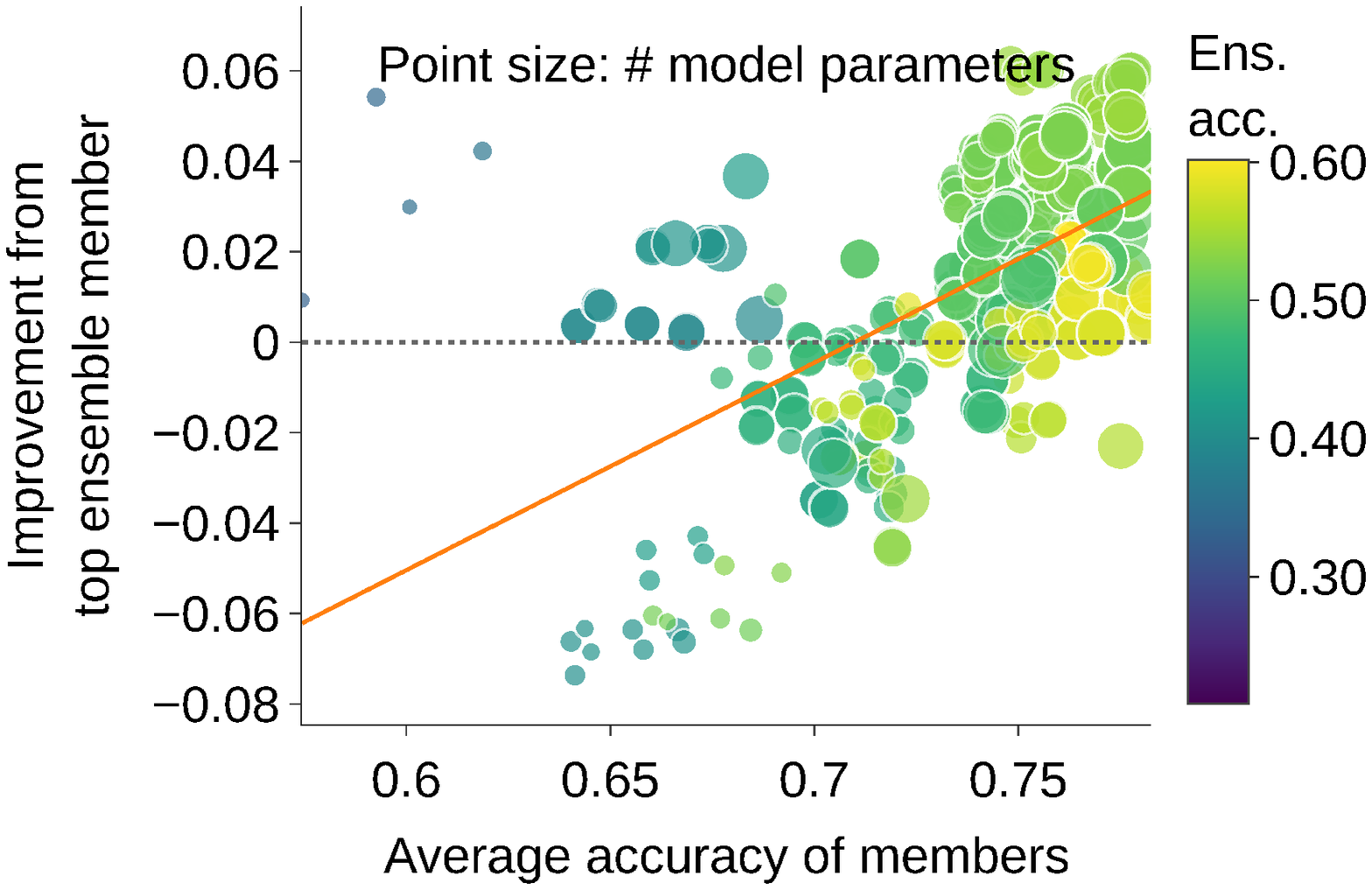}}
            \caption{Comparison of \textbf{trend lines}, i.e., correlation of improvement to three different metrics on five validation datasets (consensus: averaging). Columns: Datasets. Rows: Metrics. The attribution metric (a to e) has a much lower accuracy trade-off compared to the disagreement metric (f to j) but is higher than the average accuracy metric (k to o).
            Bigger plots to appreciate individual ensembles are presented in Section A.2 of the supplemental material.
            }
            \label{fig:heteroODD}
        \end{figure}

        We evaluate the same ensembles on five more validation datasets and verify that the observed trend in the validation dataset applies to natural corruptions.
        In addition, we compare the two diversity metrics to a simple validation accuracy metric.
        See Figure~\ref{fig:heteroODD}.

        \textbf{Observations to Figure~\ref{fig:heteroODD}:} Attribution-based diversity is better correlated as well.
        These results serve as evidence to confirm that the diversity-accuracy trade-off is better for attribution than for prediction diversity.
        However, the metric of averaging the individual accuracies of the ensemble members is more strongly correlated with the ensemble improvement in corruptions.

\begin{figure}[htp]
            \subcaptionbox{Attribution\label{fig:Autoa}}{\includegraphics[width=0.33\linewidth]{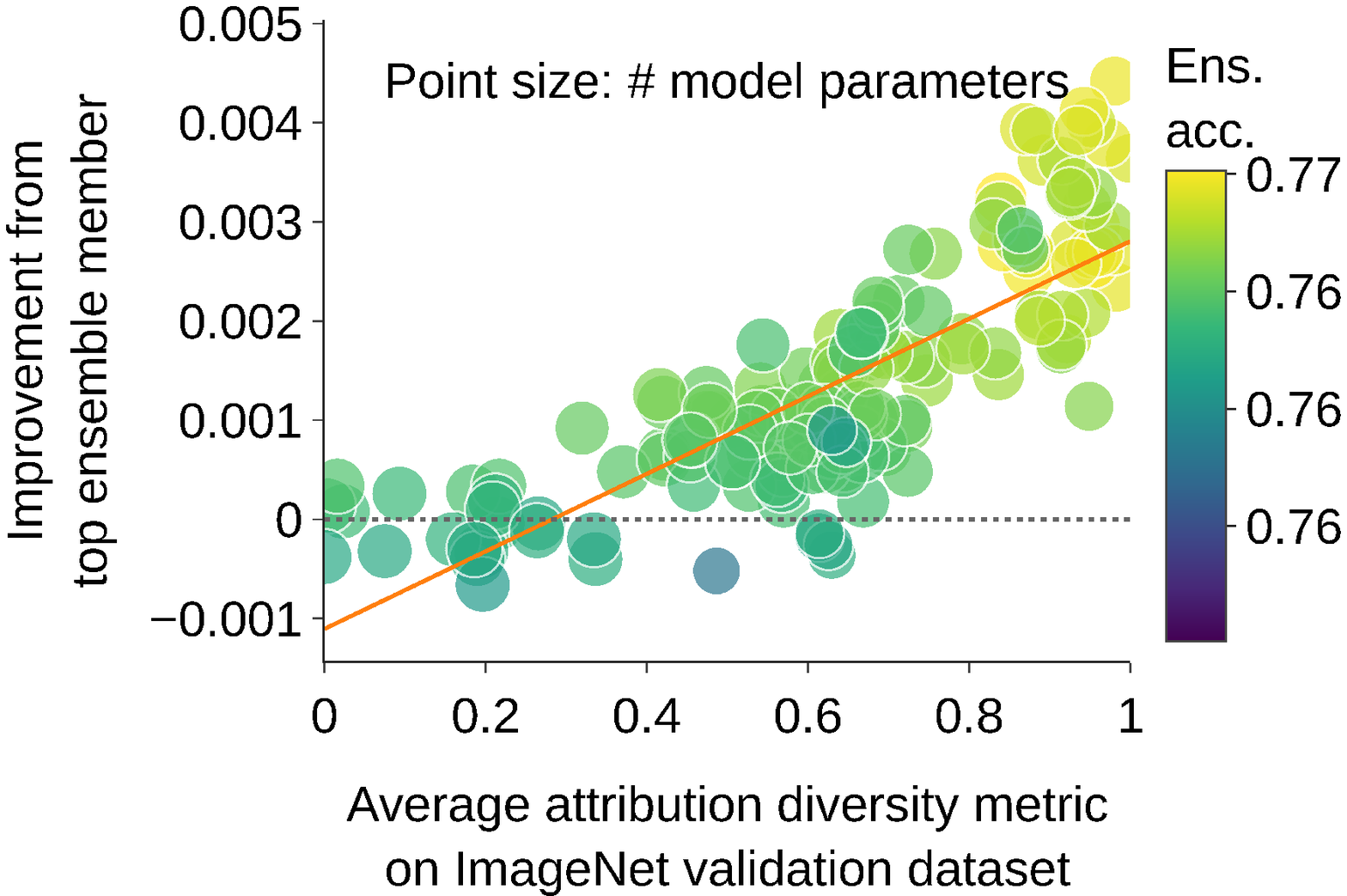}}\hspace{0em}%
            \subcaptionbox{Disagreement\label{fig:Autob}}{\includegraphics[width=0.33\linewidth]{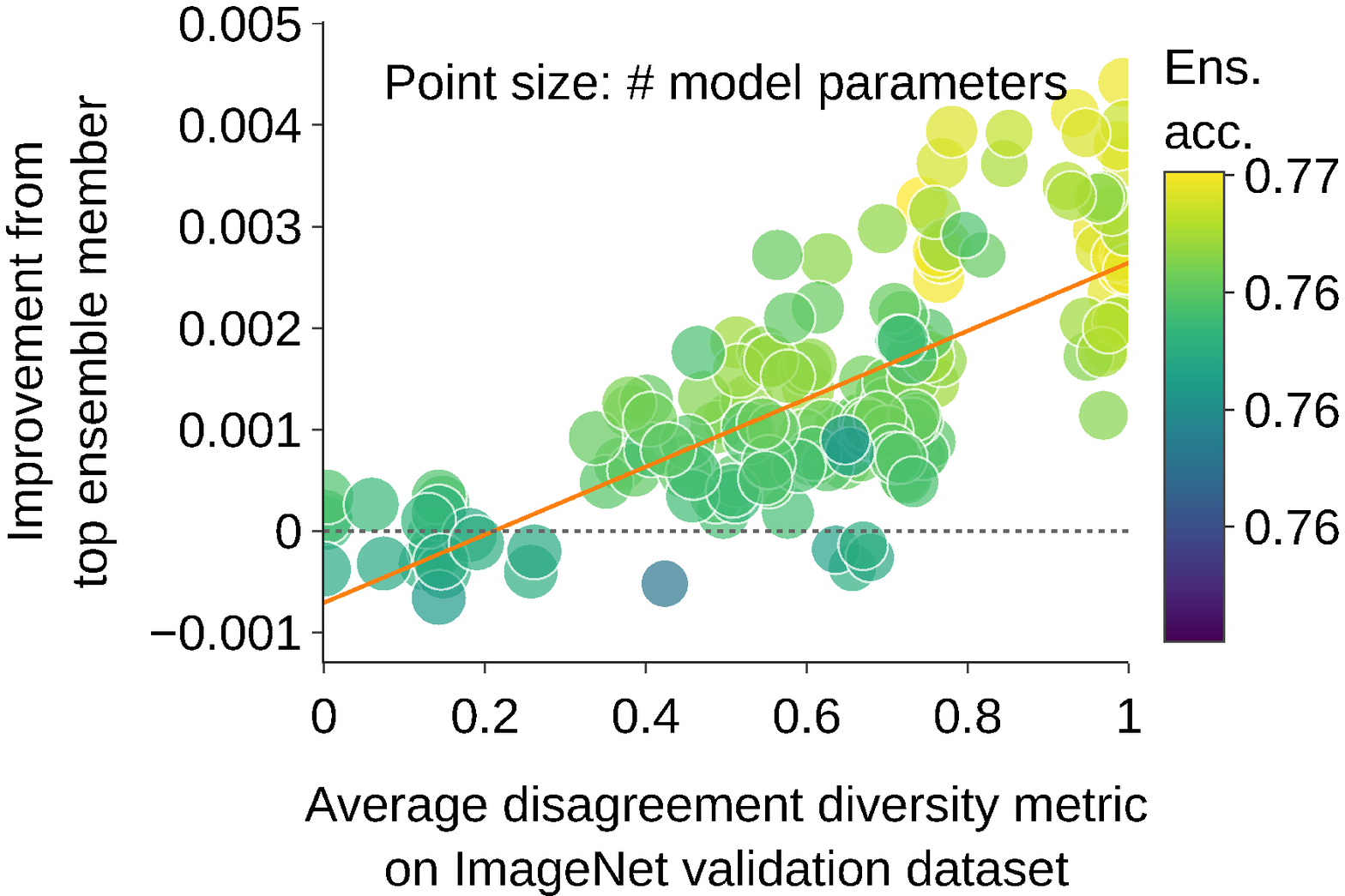}}
            \subcaptionbox{Accuracy\label{fig:Autoc}}{\includegraphics[width=0.33\linewidth]{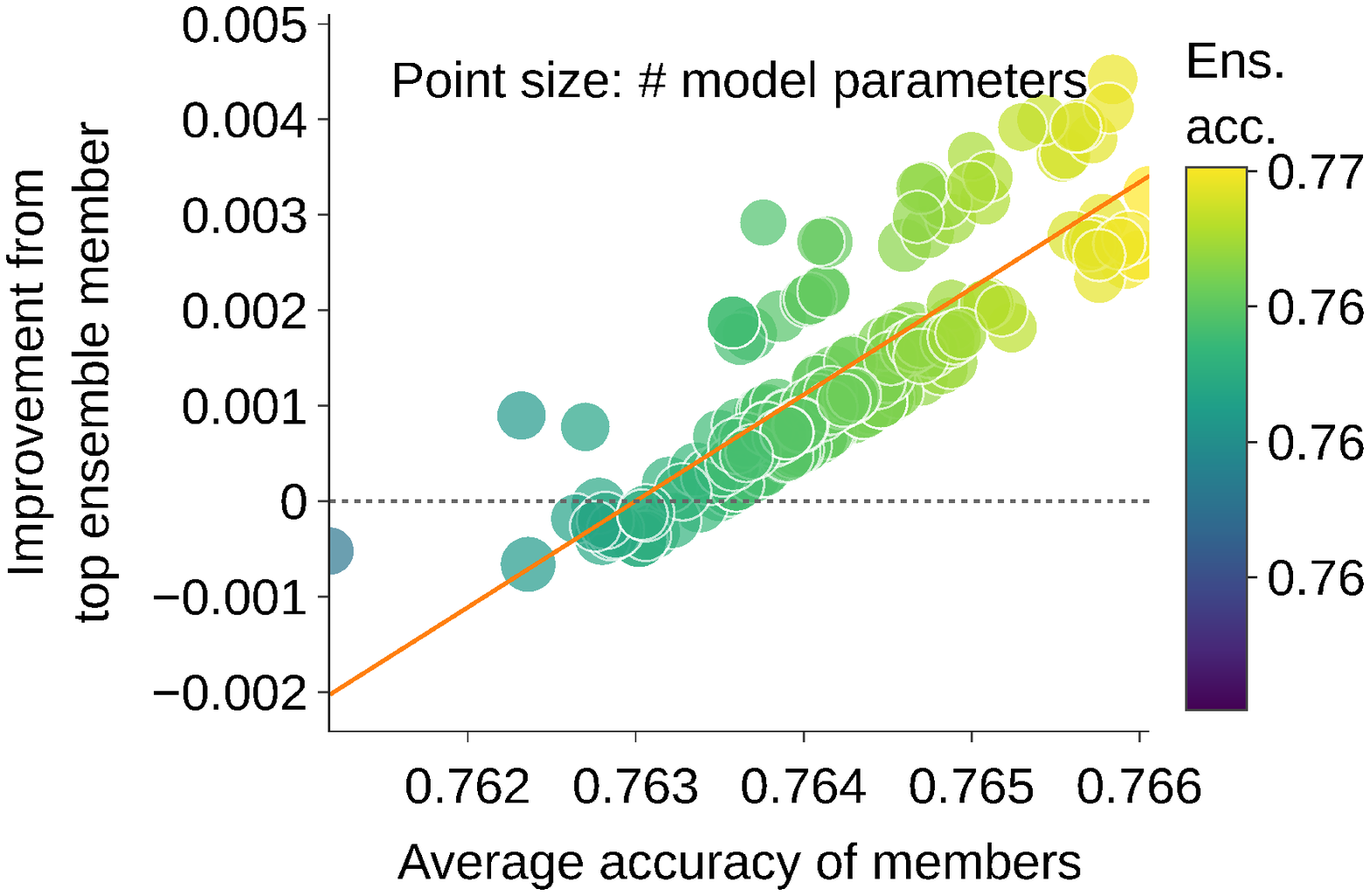}}        
            \caption{Comparison of 165 heterogeneous ensembles of \textbf{architectures automatically created with weight-sharing neural architecture search (NAS)}. 11 models with different architectures were selected to be very close in complexity, i.e., number of parameters. The correlation between the two diversity metrics is highly positive but the increment in performance is very small.} \label{fig:heteroAutoML}
        \end{figure}

        Next, Figure~\ref{fig:heteroAutoML} presents the results of the second experiment on architectures created with NAS.
        We used the open-source framework BootstrapNAS \cite{munoz2022enabling, munoz2022automated} to create a weight-sharing super-network.
        The super-network is trained from an initial ResNet50 model.
        We then sample 11 subnetworks with different configurations but similar complexity by varying the width and depth of the CNN.

        \textbf{Observations to Figure~\ref{fig:heteroAutoML}:} Although the correlations seem strong for all metrics, the actual ensemble improvement is very low, i.e., less than 0.04\%.

\begin{figure}[htbp]
            \centering
            \includegraphics[width=\linewidth]{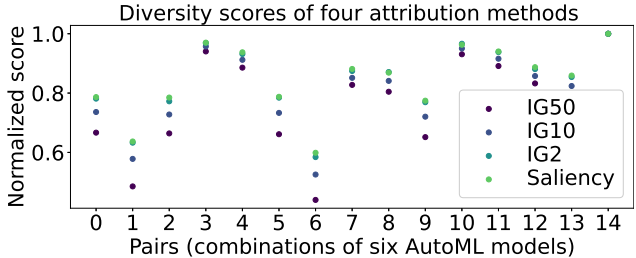}
            \caption{Comparison of Saliency and Integrated Gradients with 2, 10 \& 50 samples.
            The X-axis represents different pairs of models created with BootstrapNAS.
            The Y-axis is the normalized diversity score of each method.
            }\label{fig:attMethodsComp}
        \end{figure}

        To identify the effect of the complexity of the chosen attribution method, we evaluated the pair-wise diversity on the entire validation set on six subnetworks using \textit{Saliency} and \textit{Integrated Gradients} attribution methods using 1, 2, 10, and 50 backpropagation passes. See Figure~\ref{fig:attMethodsComp}.
        The average correlation coefficient of the normalized diversity scores of all methods is 0.998.
        Using Saliency-based attribution is then justified as it provides the lowest performance penalty as the computational overhead grows linearly with the number of backpropagation passes.

    \subsection{Enforcing diversity in homogeneous ensembles}
    \label{labelExpHomo}

\begin{figure}[htp]
            {\includegraphics[width=\linewidth]{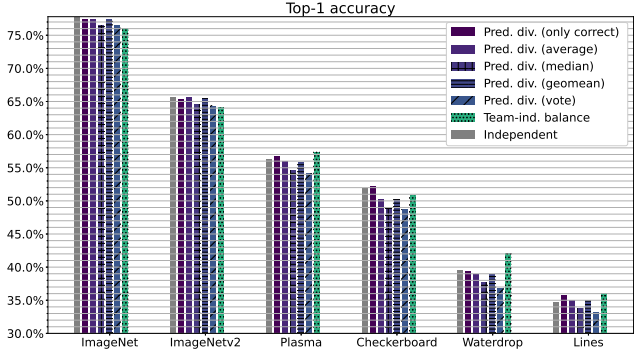}}\hspace{0em}%
            \caption{The resulting accuracy and \textbf{resiliency to natural corruptions of diversity enforcement} based on Negative Correlation Learning with different consensus mechanisms.
            Each ensemble has three members with a ResNet50 architecture.
            The heuristic of balancing the loss of the ensemble and the loss of the individual member produced the most resilient ensemble to corruptions.
            } \label{fig:gncl}
        \end{figure}

        We perform a set of training experiments to enforce diversity into the ensembles through the loss function via the Negative Correlation Learning paradigm.
        We use ResNet50 for all ensemble members, and evaluate different heuristics: \\
        a) Independently trained members using cross-entropy as loss in Equation~\ref{eq:ind}.
        Four different consensus approaches in GNCL using Equation~\ref{eq:gncl_i}:
        b) average,
        c) median,
        d) geometric mean,
        e) majority vote,
        f) GNCL and averaging consensus but masking the penalty term for incorrect classifications, i.e., $ (h^{i} \neq y) \Rightarrow (\lambda=0)$, and
        g) Balancing a loss function between the team and individual members (Equation~\ref{eq:balanced}).
        The optimization method in all cases was AdaBelief~\cite{ZhuangTDTDPD20} for 100 epochs with a learning rate of 1e-3 decaying 10\% every 30 epochs, epsilon of 1e-8, betas: (0.9,0.999), batch size of 64 and a $\lambda$ factor of 0.2.
        In ImageNet classification, we empirically observed that bigger $\lambda$ values in Equation~\ref{eq:gncl_i} fail to learn.
        Results for the six heuristics are presented in Figure~\ref{fig:gncl}.
		
        \textbf{Observations to Figure~\ref{fig:gncl}:} Explicit enforcement of prediction diversity does not result in improved resilience. However the balanced loss (Eq.~\ref{eq:balanced}) provides a significant advantage in 3 out of 4 natural corruptions.

    \subsubsection{First attempt at enforcing attribution diversity}

        We perform a first attempt to enforce attribution diversity using the loss of Equation~\ref{eq:attDivEnf} and the same optimization parameters used in GNCL.
        The computational overhead to calculate the attributions is 2x using the Saliency method.
        Empirically, we tried five different lambda weights values: \{10, 1,0.1,0.01,0.001\} but found training instabilities.
        The smallest $\lambda$ value resulted in convergence up to epoch 21 for 63.7\% top1 accuracy.
        We believe that the penalty term of Eq.~\ref{eq:attDivEnf} is in conflict with the original loss and it would be more appropriate to investigate a better penalty term than to optimize this hyper-parameter in future work.

    \subsubsection{Diversity of NCL ensembles}

\begin{figure}
		\includegraphics[width=\linewidth]{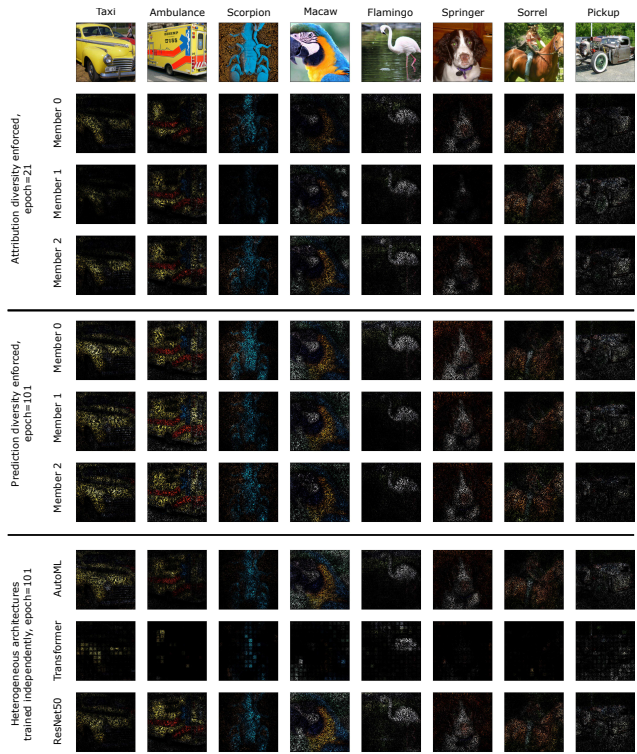}
		\caption{\textbf{Attribution map diversity} of different diversity-inducing techniques on 8 ImageNet val. dataset samples.} \label{fig:attmapsAll}
	\end{figure}

\begin{figure}[htp]
            \centering
            \subcaptionbox{\centering Attribution diversity enf.\label{fig:ckaa}}{\includegraphics[width=0.26\linewidth]{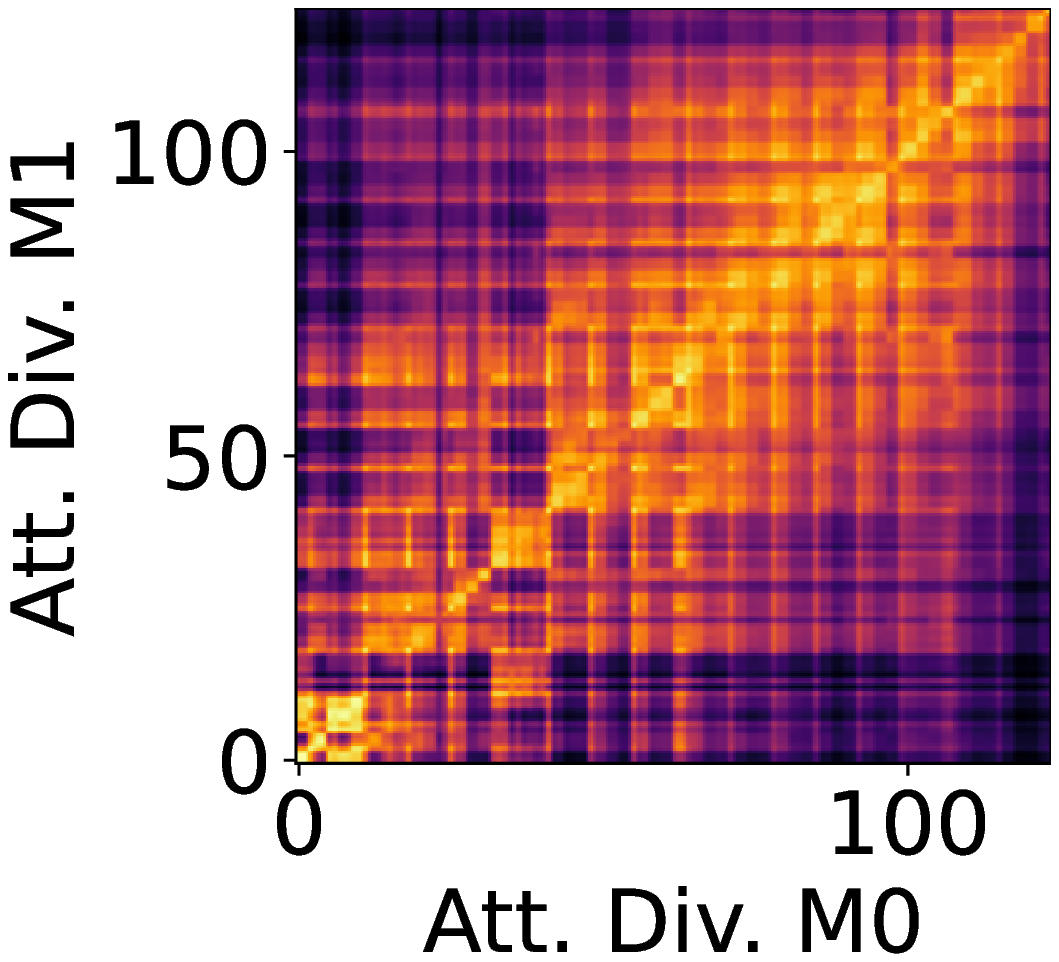}}\hspace{1em}%
            \subcaptionbox{\centering Prediction diversity enf.\label{fig:ckab}}{\includegraphics[width=0.26\linewidth]{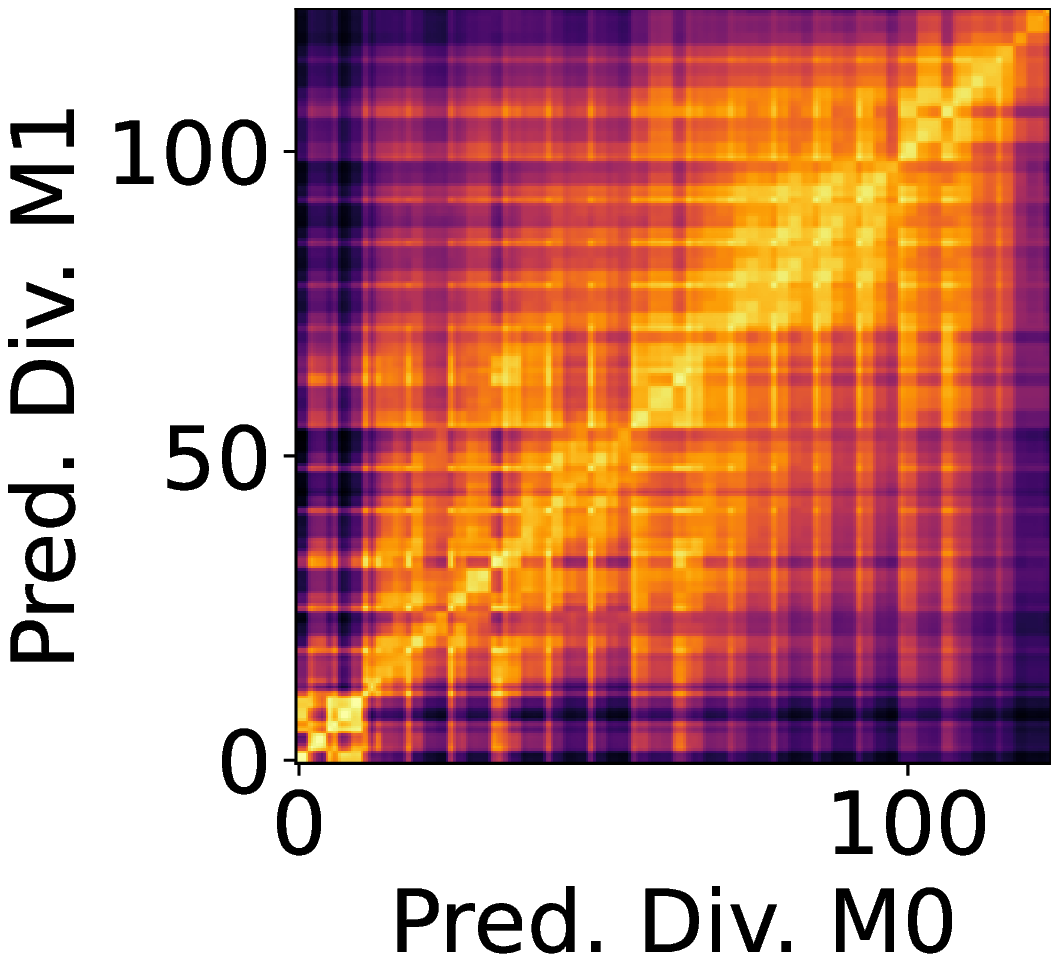}}\hspace{1em}%
            \subcaptionbox{\centering Heterog. architectures\label{fig:ckac}}{\includegraphics[width=0.26\linewidth]{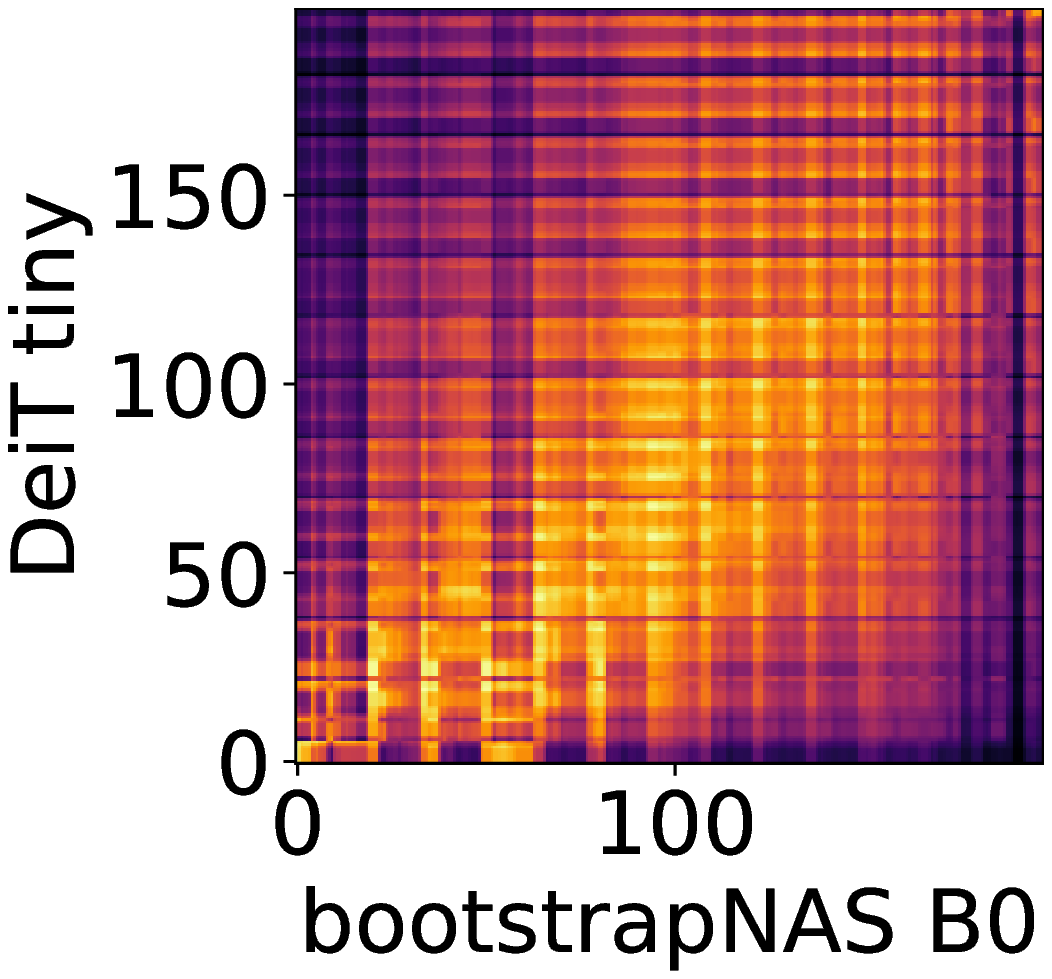}}\hspace{1em}%
            {\includegraphics[width=0.08\linewidth]{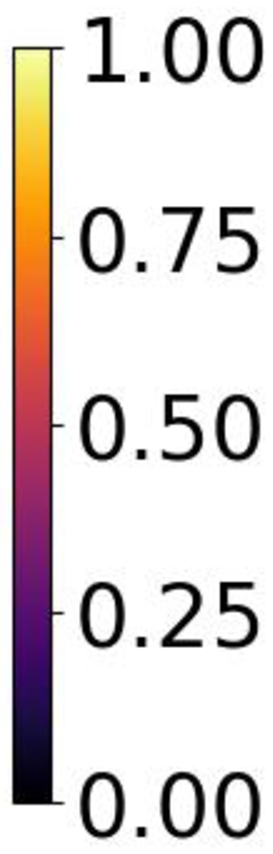}}
            \caption{Centered Kernel Alignment maps to visualize the resulting \textbf{similarity of model layers} on different diversity enforcing techniques.} \label{fig:cka}
        \end{figure}

        Figures~\ref{fig:attmapsAll},~\ref{fig:cka} together with Table~\ref{tab:shanon} show three types of diversity (attribution, prediction, and intermediate representation) for three models created through independently created heterogeneous architectures, prediction diversity enforcement and attribution diversity enforcement with the following top1 accuracies on the ImageNet validation dataset: 78.2\%, 76.1\%, and 63.7\%.

        \textbf{Prediction diversity.}
        In Table~\ref{tab:shanon}, the Shannon equitability index metric (Eq.~\ref{eq:shannon})  is shown for correctly and incorrectly classified samples for three ensembles: attribution diversity (Eq.~\ref{eq:attDivEnf}), prediction diversity (Eq.~\ref{eq:disagreement}) and heterogeneous architectures on all six datasets.
        The heterogeneous ensemble produces more diverse predictions in general.
		
	\begin{table}
            \footnotesize
            \caption{\textbf{Diversity of predictions} from all members of three ensembles as measured by the Shanon equitability index $H$. $H_{corr}$ and $H_{inco}$ indicate the metric computed on all samples that were correctly or incorrectly classified. The six subcolumns correspond to the six validation datasets.}\label{tab:shanon}
            \resizebox{\linewidth}{!}{%
            \begin{tabular}{|l|c|c|c|c|c|c|c|c|c|c|c|c|}
            \hline
             &\multicolumn{6}{c|}{$H_{corr}$}& 
             \multicolumn{6}{c|}{$H_{inco}$}\\
            &IN&I2&WD&LI&PL&CB& IN&I2&WD&LI&PL&CB\\
            \hline
            Att. div.&\underline{0.13}&0.16&0.29&0.32&0.23&0.23&0.46&0.49&0.62&0.60&0.57&0.58\\
            Pred. div.&0.10&0.13&0.29&0.31&0.21&0.23&0.46&0.49&0.67&0.69&0.59&0.63\\    Hetero.&0.12&\underline{0.17}&\underline{0.35}&\underline{0.37}&\underline{0.25}&\underline{0.29}&\underline{0.51}&\underline{0.55}&\underline{0.74}&\underline{0.77}&\underline{0.67}&\underline{0.71}\\
            \hline
            \end{tabular}
            }
        \end{table}

        \textbf{Attribution diversity.}
        We present a few resulting attribution maps in Figure~\ref{fig:attmapsAll} for the NCL-based prediction-diversity enforcement, attribution-diversity enforcement (at epoch 21), and independently trained architectures.

        \textbf{Observations to Figure~\ref{fig:attmapsAll}:} Independently trained heterogeneous architectures and attribution-diversity enforcement produce more diverse attribution maps than homogeneous models trained to have diverse prediction outcomes.

        \textbf{Representation diversity.}
        In Figure~\ref{fig:cka}, we investigate the resulting diversity/similarity of the internal layers via CKA (Equation~\ref{eq:cka}) of two ensemble members for three different diversity enforcing techniques.

        \textbf{Observations to Figure~\ref{fig:cka}:} The CKA visualization reflects that the enforcement of attribution diversity produces less similarity in the layers than output diversity enforcement or by independent heterogeneous architectures.

\section{Discussion and conclusions}
\label{labelDiscussion}

    In this section, we discuss and interpret the results of Section~\ref{labelExperiments} and summarize the research questions' answers.

    \subsection{Answers to research questions}
        \refreq{accRes}: In our experiments, it is observed that model architecture is more important to resiliency than model accuracy or size.
        On \refreq{att}, we consistently observed that attribution-based diversity is more positively correlated with accuracy than prediction-based disagreement diversity.
        Answering \refreq{bestEnf}, balancing the loss of the individual members and the ensemble provided a significant advantage in 3 out of 4 natural corruptions when compared to the prediction diversity enforcement variants.
        \refreq{divNCL}: Prediction diversity was higher for heterogeneous architectures trained independently than NCL on prediction or attribution diversity.
        Attribution diversity is significantly lower when enforcing prediction diversity compared to heterogeneous architectures trained independently.
        Activation diversity is low at the last layers for both prediction and attribution diversity enforcement, while for heterogeneous architectures trained independently, the middle layers showed less diversity.
        
    \subsection{Results discussions}

        \textbf{Advantage of Transformers vs CNNs:} The superiority of the transformer architecture in terms of resilience against natural corruptions could be attributed to their capability to pay attention globally instead of locally as CNNs do, and thus they may be able to construct more useful intermediate representations that suffer less from perturbations.

        \textbf{NAS ensembles in our experiments are not diverse enough:} Only small improvements are obtained from ensembles created from sub-networks.
        These are jointly trained as part of a single weight-sharing super-network and thus have little diversity. Larger search/design spaces should be explored in future work.

        \textbf{Balancing member vs ensemble accuracy:} Explicitly enforcing prediction diversity is outperformed by implicit enforcement through the balance of ensemble and individual accuracy (Equation~\ref{eq:balanced}).
        This could be due to the use of two less conflicting objectives than prediction diversity.

        \textbf{Diversity-accuracy trade-off improvement:} In contrast to the disagreement metric, attribution diversity does not require models to deviate from a correct prediction.
        The correlation is however not very strong as models tend to find similar features for prediction, and implicit attribution deviations may be the product of imperfect learning.
        The enforcement of attribution diversity with the proposed loss proved however to be insufficient and better heuristics need to be explored.
        
        \textbf{Diverse architectures produced more diversity than NCL-based methods}
        An ensemble selected from a combination of independently trained heterogeneous architectures and training approaches resulted in higher levels of diversity in prediction, attribution, and intermediate layers.
        However, mixing different architectures does not consistently produce good ensembles as observed in the many ensembles under the zero line in Figure~\ref{fig:heteroODD}, as a good model may not benefit from ensembling with a less good one, as their common modes are not negatively correlated.

    \subsection{Conclusions and next steps}

        In this study, we explored different approaches to measure and enforce diversity in ensembles and evaluated their impact on natural data corruption resiliency.
        The key takeaways are:
        1) model architecture is more important for resiliency than model size or model accuracy,
        2) attribution-based diversity is less negatively correlated to the ensemble accuracy than prediction-based diversity,
        3) a balanced loss function of individual and ensemble accuracy creates more resilient ensembles for image natural corruptions, and
        4) architecture diversity produces more diversity in all explored diversity metrics: predictions, attributions, and activations.
        
        In addition, other valuable findings are:
        a) Saliency attribution can be sufficient to measure input attribution diversity,
        b) Ensembles created from models of similar complexity that were discovered by weight-sharing Neural Architecture Search for our experiments barely provide any accuracy improvement, and
        c) Enforcing attribution-based diversity during training through a variance-based penalty term is not stable and needs further research.
        
        In future work, several experiments could be done to understand the
        complexity-resiliency trade-off, e.g., through knowledge distillation,
        improved heuristics to enforce attribution diversity, and compare diversity approaches in tasks beyond image classification.

%% file: suplemental.tex
\onecolumn

\renewcommand{\thesection}{\Alph{section}}
\renewcommand{\thetable}{\roman{table}}
\renewcommand{\thefigure}{\Alph{figure}}

\setcounter{section}{0}
\setcounter{table}{0}
\setcounter{figure}{0}

\section{Supplemental material}

\subsection{Training parameters for heterogeneous architectures}

Table~\ref{tab:tParams} shows the architecture and optimization hyper-parameters used for training the  models used in our experiments.

\begin{table}[H]
\centering
\resizebox{\linewidth}{!}{
\begin{tabular}{|c|c|c|c|c|c|c|}
\hline
id & Architecture & Optimizer & Parameters & Scheduler & Epochs &
BatchSize\\
\hline
ResNext50\_32\_2&ResNext50 cardinality=32, blockWidth=2 layers=[3;4;6;3], dropout=0.2 & SGD & lr=0.1 decay=0.0001, momentum=0.9 & Step: $\gamma$=0.1, epochs=30 & 100 & 32\\
\hline
ResNext50\_32\_4l&ResNext50 cardinality=32, blockWidth=4 layers=[2;2;3;2], dropout=0.2 & SGD & lr=0.1 decay=0.0001, momentum=0.9 & Step: $\gamma$=0.1, epochs=30 & 100 & 32\\
\hline
ResNext50\_16\_4&ResNext50 cardinality=16, blockWidth=4 layers=[3;4;6;3], dropout=0.2 & SGD & lr=0.1 decay=0.0001, momentum=0.9& Step: $\gamma$=0.1, epochs=30& 100 & 32\\
\hline
MNASNET\_1&MNASNET ratio=1, dropout=0.2 & SGD & lr=0.1 decay=0.0001, momentum=0.9& Step: $\gamma$=0.1, epochs=30& 100 & 32\\
\hline
MNASNET\_1p&MNASNET ratio=1, dropout=0.2 & RMSprop & lr=0.256, decay=0.9, momentum=0.9& Step: $\gamma$=0.97, epochs=2.4& 100 & 32\\
\hline
Squeezenet\_512 & Squeezenet version=1.1 & SGD & lr=0.01 decay=0.0002, momentum=0.9& Step: $\gamma$=0.1, epochs=30& 100 & 512\\
\hline
Squeezenet & Squeezenet version=1.1 & SGD & lr=0.01 decay=0.0002, momentum=0.9& Step: $\gamma$=0.1, epochs=30& 100 & 128\\
\hline
\multirow{2}{*}{bootstrapNAS-B\_0} & ResNet50 depth=[0, 0, 0, 0, 1], width=[0, 0, 0, 2, 2, 2]&\multirow{2}{*}{SGD} &\multirow{2}{*}{See Listing~\ref{lst:progShr}} &\multirow{2}{*}- &\multirow{2}{*}-&\multirow{2}{*}-\\
& expansion=[0.2, 0.2, 0.2, 0.25, 0.2, 0.25, 0.25, 0.25,0.2, 0.25, 0.25, 0.25] & & & & &\\
\hline
\multirow{2}{*}{bootstrapNAS-B\_1} & ResNet50 d=[0, 0, 0, 0, 1], w=[0, 0, 0, 2, 2, 2]&\multirow{2}{*}{SGD} &\multirow{2}{*}{See Listing~\ref{lst:progShr}} &\multirow{2}{*}- &\multirow{2}{*}- & \multirow{2}{*}-\\
& expansion=[0.25, 0.2, 0.25, 0.25, 0.2, 0.25, 0.25, 0.25, 0.2, 0.25, 0.25, 0.25] & & & & &\\
\hline
deit\_tiny\_p16\_224 & DeiT size=tiny(3 heads), patchSize=16, embedding=192& AdamW & lr=0.0005 & Cosine & 300 & 256\\
\hline
\multirow{2}{*}{dino\_deit\_tiny1} & DINO DeiT size=tiny, patchSize=16, embedding=192, & \multirow{2}{*}{AdamW} & \multirow{2}{*}{lr=0.0005} & \multirow{2}{*}{Cosine} & Backbone=300 & \multirow{2}{*}{256}\\
&seed=0, localCropsScale=\{0.05,0.2\}, globalCropsScale=\{0.4, 1.\}&&&& Classifier=100&\\
\hline
\multirow{2}{*}{dino\_deit\_tiny2} & DINO DeiT size=tiny, patchSize=16, embedding=192, & \multirow{2}{*}{AdamW} & \multirow{2}{*}{lr=0.0005} & \multirow{2}{*}{Cosine} & Backbone=300 & \multirow{2}{*}{256}\\
&seed=7, localCropsScale=\{0.05,0.2\}, globalCropsScale=\{0.4, 1.\}&&&& Classifier=100&\\
\hline
\multirow{2}{*}{dino\_resnet1} & DINO ResNet50 backboneSeed=7 localCropsScale=\{0.05,0.14\} & \multirow{2}{*}{SGD} & \multirow{2}{*}{decay=0.0001, lr=0.03} & \multirow{2}{*}{-} & Backbone=300 & \multirow{2}{*}{256}\\
&globalCropsScale=\{0.14, 1.\}, classifierSeed=5&&&& Classifier=100&\\
\hline
\multirow{2}{*}{dino\_resnet2} & DINO ResNet50 backboneSeed=7 localCropsScale=\{0.05,0.14\} & \multirow{2}{*}{SGD} & \multirow{2}{*}{decay=0.0001, lr=0.03} & \multirow{2}{*}{-} & Backbone=300 & \multirow{2}{*}{256}\\
&globalCropsScale=\{0.14, 1.\}, classifierSeed=15&&&& Classifier=100&\\
\hline
\end{tabular}
            }
\caption{Training architectures and parameters used to create ensembles of heterogeneous architectures}
\label{tab:tParams}
\end{table}

Listing~\ref{lst:progShr} shows an example configuration file to create a super-network using a pre-trained model from Torchvision.
The configuration parameters are used by BootstrapNAS in the Neural Network Compression Framework (NNCF) and specify the size and elasticity of the super-network.  Once the super-network has been trained, the user can extract models of different sizes and performances.

{\setstretch{0.5}\tiny
\begin{lstlisting}[label=lst:progShr, linewidth=\columnwidth, language=json, caption=Super-network configuration example for NNCF's BootstrapNAS]
# Insert here model and dataset fields
# Insert here optimizer fields
"bootstrapNAS":{
    "training": {
        "algorithm":"progressive_shrinking",   
        "progressivity_of_elasticity": ["depth", "width"], 
        "batchnorm_adaptation": {
            "num_bn_adaptation_samples": 1500},
    "schedule": { 
        "list_stage_descriptions": [
        {"train_dims": ["depth"], "epochs": 25, 
        "depth_indicator": 1, "init_lr": 2.5e-6, 
        "epochs_lr": 25},
        {"train_dims": ["depth"], "epochs": 40, "depth_indicator": 2, "init_lr": 2.5e-6, "epochs_lr": 40},
        {"train_dims": ["depth", "width"], "epochs": 50, "depth_indicator": 2, "reorg_weights": true, "width_indicator": 2, "bn_adapt": true, "init_lr": 2.5e-6, "epochs_lr": 50},
        {"train_dims": ["depth", "width"], "epochs": 50, "depth_indicator": 2, "reorg_weights": true, "width_indicator": 3, "bn_adapt": true, "init_lr": 2.5e-6, "epochs_lr": 50}
        ]
    }, 
    "elasticity": {            
        "available_elasticity_dims": ["width", "depth"],
        "width": {
        "max_num_widths": 3,
        "min_width": 32,
        "width_step": 32, 
        "width_multipliers": [1, 0.80, 0.60]
        }
    }    
},
    "search": {
        "algorithm": "NSGA2",
        "num_evals": 1000,
        "population": 50,
        "ref_acc": 93.65
    }
}
\end{lstlisting}
}

The parameters shown in Table~\ref{tab:nasParam} indicate the configuration of the subnetworks extracted from the super-network in our
experiments. We used a previous version of BootstrapNAS in our experiments, which extends Once-for-all (OFA) super-networks from Cai et al. [4] and follows its conventions to describe the search space. In newer versions of BootstrapNAS, expansion ratios are handled by an elastic width handler, and the search space description follows a different convention. 

\begin{table}[H]
\centering
\resizebox{\linewidth}{!}{
\begin{tabular}{|c|c|}
\hline
id & Subnetwork Configurations \\\hline
B\_0 & depth: [0, 0, 0, 0, 1], expansion: [0.2, 0.2, 0.2, 0.25, 0.2, 0.25, 0.25, 0.25, 0.2, 0.25, 0.25, 0.25], width: [0, 0, 0, 2, 2, 2], \\\hline
nB\_0 & depth: [0, 0, 0, 0, 1], expansion: [0.2, 0.2, 0.2, 0.25, 0.2, 0.25, 0.25, 0.25, 0.2, 0.25, 0.25, 0.25], width : [0, 0, 0, 2, 2, 2]\\\hline
nB\_1 & depth: [0, 0, 0, 0, 1], expansion:  [0.25, 0.2, 0.25, 0.25, 0.2, 0.25, 0.25, 0.25, 0.2, 0.25, 0.25, 0.25], width : [0, 0, 0, 2, 2, 2]\\\hline
nB\_2 & depth: [0, 0, 0, 0, 1], expansion:  [0.25, 0.2, 0.25, 0.2, 0.2, 0.25, 0.25, 0.25, 0.2, 0.25, 0.25, 0.25], width : [0, 0, 0, 2, 2, 2]\\\hline
nB\_3 & depth: [0, 0, 0, 0, 1], expansion:  [0.25, 0.2, 0.25, 0.25, 0.2, 0.2, 0.25, 0.25, 0.2, 0.25, 0.25, 0.25], width : [0, 0, 0, 2, 2, 2]\\\hline
dB\_1a & depth: [0, 0, 1, 0, 0], expansion: [0.25, 0.2, 0.25, 0.25, 0.2, 0.25, 0.25, 0.25, 0.2, 0.25, 0.25, 0.25], width : [0, 0, 0, 2, 2, 2]\\\hline
dB\_1b & depth: [1, 0, 0, 0, 0], expansion: [0.25, 0.2, 0.25, 0.25, 0.2, 0.25, 0.25, 0.25, 0.2, 0.25, 0.25, 0.25], width : [0, 0, 0, 2, 2, 2],\\\hline
wB\_1a & depth: [0, 0, 0, 0, 1], expansion: [0.25, 0.2, 0.25, 0.25, 0.2, 0.25, 0.25, 0.25, 0.2, 0.25, 0.25, 0.25], width : [0, 0, 1, 1, 2, 2],\\\hline
wB\_1b & depth: [0, 0, 0, 0, 1], expansion: [0.25, 0.2, 0.25, 0.25, 0.2, 0.25, 0.25, 0.25, 0.2, 0.25, 0.25, 0.25], width : [0, 1, 1, 1, 1, 2],\\\hline
wB\_1c & depth: [0, 0, 0, 0, 1], expansion: [0.25, 0.2, 0.25, 0.25, 0.2, 0.25, 0.25, 0.25, 0.2, 0.25, 0.25, 0.25], width : [1, 1, 1, 1, 1, 1],\\\hline
    \end{tabular}
    }
    \caption{Configuration of subnetworks extracted for the creation of ensembles of heterogeneous architectures}
    \label{tab:nasParam}
\end{table}

Table~\ref{tab:trainingTrans} shows the transformations used during training for all individual models.

    \begin{table}[H]
        \centering
        \begin{tabular}{|c|c|}
            \hline
            Transform & Parameters\\
             \hline
             RandomResizedCrop & size=(224, 224), scale=(0.08, 1.0), ratio=(0.75, 1.3333), interpolation=bilinear \\
             \hline
             RandomHorizontalFlip & p=0.5 \\
             \hline
             Normalize & mean=[0.485, 0.456, 0.406], std=[0.229, 0.224, 0.225]\\
             \hline
        \end{tabular}
        \caption{Training data set transforms used for the training of all models}
        \label{tab:trainingTrans}
    \end{table}

Figure~\ref{tab:indAccNAS} shows the final top1 accuracy scores of each model defined by Table~\ref{tab:tParams}.

\begin{figure}[htbp]
    \begin{subtable}[t]{0.45\textwidth}
        \centering
        \begin{tabular}{|c|c|}
            \hline
            Model id & Top 1\% accuracy \\
            \hline
            ResNext50\_32\_2 & 75.198  \\
            ResNext50\_32\_4l & 75.072 \\
            ResNext50\_16\_4 & 75.432 \\
            MNASNET\_1 & 54.408 \\
            MNASNET\_1p & 54.0 \\
            Squeezenet\_512 & 41.918 \\
            Squeezenet & 56.558 \\
            bootstrapNAS-B\_0 & 76.342 \\
            bootstrapNAS-B\_1 & 76.282 \\
            deit\_tiny\_p16\_224 & 71.654 \\
            dino\_deit\_tiny1 & 67.932 \\
            dino\_deit\_tiny2 & 67.52 \\
            dino\_resnet1 & 67.236 \\
            dino\_resnet2 & 67.216 \\
            \hline
        \end{tabular}
    \end{subtable}
    \quad
    \begin{subtable}[t]{0.45\textwidth}
        \centering
        \begin{tabular}{|c|c|}
            \hline
            Model id & Top 1\% accuracy \\
            \hline
            B\_0 & 76.342 \\
            B\_1 & 76.282 \\
            nB\_0 & 76.301\\
            nB\_1 & 76.318\\
            nB\_2 & 76.191\\
            nB\_3 & 76.142\\
            dB\_1a & 76.138\\
            dB\_1b & 76.084\\
            wB\_1a & 76.170\\
            wB\_1b & 75.804\\
            wB\_1c & 75.852\\
            \hline
        \end{tabular}
        
    \end{subtable}
    \caption{Individual accuracies of individual trained models on ImageNet validation dataset}
        \label{tab:indAccNAS}
\end{figure}

\newpage

\subsection{Comparison of prediction-based disagreement, input attribution diversity and average accuracy metrics in ensembles of heterogeneous architectures}

    \begin{figure}[htp]
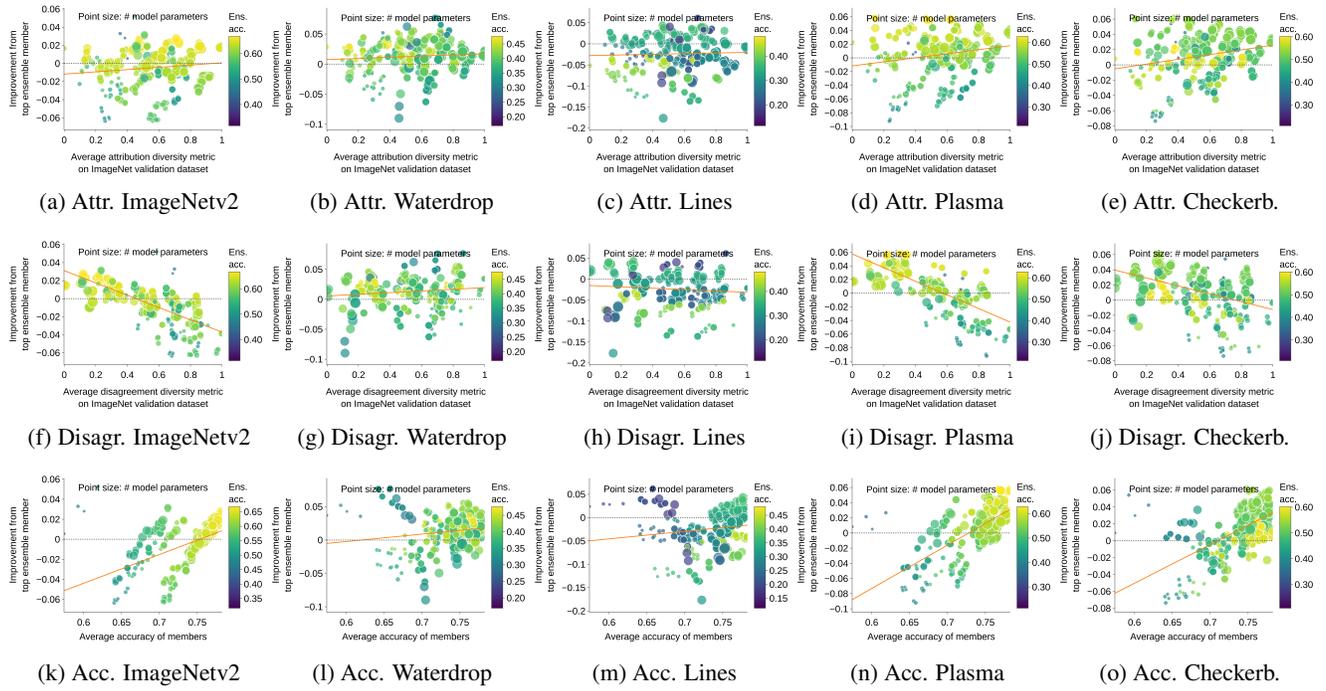

        \subcaptionbox{\centering Attr. ImageNetv2\label{figB:odda}}{\includegraphics[width=0.2\linewidth]{figures/fig_hetero3_exp_linear_combination_DIFF_imagenetv2.eps}}%
        \hspace{0em}%
        \subcaptionbox{\centering Attr. Waterdrop\label{figB:oddb}}{\includegraphics[width=0.2\linewidth]{figures/fig_hetero3_exp_linear_combination_DIFF_waterdrop_7.eps}}%
        \hspace{0em}%
        \subcaptionbox{\centering Attr. Lines\label{figB:oddc}}{\includegraphics[width=0.2\linewidth]{figures/fig_hetero3_exp_linear_combination_DIFF_lines.eps}}%
        \hspace{0em}%
        \subcaptionbox{\centering Attr. Plasma\label{figB:oddd}}{\includegraphics[width=0.2\linewidth]{figures/fig_hetero3_exp_linear_combination_DIFF_plasma_noise.eps}}%
        \hspace{0em}%
        \subcaptionbox{\centering Attr. Checkerb.\label{figB:odde}}{\includegraphics[width=0.2\linewidth]{figures/fig_hetero3_exp_linear_combination_DIFF_checkboard.eps}}%
        \hspace{0em}%
        \subcaptionbox{\centering Disagr. ImageNetv2\label{figB:oddf}}{\includegraphics[width=0.2\linewidth]{figures/fig_hetero3_exp_linear_combination_OUT_DIS_imagenetv2.eps}}%
        \hspace{0em}%
        \subcaptionbox{\centering Disagr. Waterdrop\label{figB:oddg}}{\includegraphics[width=0.2\linewidth]{figures/fig_hetero3_exp_linear_combination_OUT_DIS_waterdrop_7.eps}}%
        \hspace{0em}%
        \subcaptionbox{\centering Disagr. Lines\label{figB:oddh}}{\includegraphics[width=0.2\linewidth]{figures/fig_hetero3_exp_linear_combination_OUT_DIS_lines.eps}}%
        \hspace{0em}%
        \subcaptionbox{\centering Disagr. Plasma\label{figB:oddi}}{\includegraphics[width=0.2\linewidth]{figures/fig_hetero3_exp_linear_combination_OUT_DIS_plasma_noise.eps}}%
        \hspace{0em}%
        \subcaptionbox{\centering Disagr. Checkerb.\label{figB:oddj}}{\includegraphics[width=0.2\linewidth]{figures/fig_hetero3_exp_linear_combination_OUT_DIS_checkboard.eps}}%
        \hspace{0em}%
        \subcaptionbox{\centering Acc. ImageNetv2\label{figB:oddk}}{\includegraphics[width=0.2\linewidth]{figures/fig_hetero3_exp_justAcc_imagenetv2.eps}}%
        \hspace{0em}%
        \subcaptionbox{\centering Acc. Waterdrop\label{figB:oddl}}{\includegraphics[width=0.2\linewidth]{figures/fig_hetero3_exp_justAcc_waterdrop_7.eps}}%
        \hspace{0em}%
        \subcaptionbox{\centering Acc. Lines\label{figB:oddm}}{\includegraphics[width=0.2\linewidth]{figures/fig_hetero3_exp_justAcc_lines.eps}}%
        \hspace{0em}%
        \subcaptionbox{\centering Acc. Plasma\label{figB:oddn}}{\includegraphics[width=0.2\linewidth]{figures/fig_hetero3_exp_justAcc_plasma_noise.eps}}%
        \hspace{0em}%
        \subcaptionbox{\centering Acc. Checkerb.\label{figB:oddo}}{\includegraphics[width=0.2\linewidth]{figures/fig_hetero3_exp_justAcc_checkboard.eps}}
        \caption{Comparison of \textbf{improvement correlation of three different metrics on five validation datasets} using averaging as consensus mechanism. Columns: Datasets. Rows: Metrics.}
        \label{figB:supHeteroODD}
        \end{figure}